\documentclass[lettersize,journal]{IEEEtran}
\usepackage{threeparttable}
\usepackage{multirow}
\usepackage{amsmath,amsfonts}
\usepackage{algorithmic}
\usepackage{algorithm}
\usepackage{array}
\usepackage[caption=false,font=normalsize,labelfont=sf,textfont=sf]{subfig}
\usepackage{textcomp}
\usepackage{stfloats}
\usepackage{url}
\usepackage{verbatim}
\usepackage{graphicx}
\usepackage{cite}
\begin{document}

\title{A Survey on Segment Anything Model (SAM): Vision Foundation Model Meets Prompt Engineering}
\author{Chaoning Zhang, \IEEEmembership{Senior Member,~IEEE}, Joseph Cho, Fachrina Dewi Puspitasari,  Sheng Zheng, Chenghao Li, Yu Qiao, Taegoo Kang, Xinru Shan, Chenshuang Zhang, Caiyan Qin, Francois Rameau, Lik-Hang Lee, Sung-Ho Bae, Choong Seon Hong  \IEEEmembership{Fellow,~IEEE} 
\thanks{Chaoning Zhang, Joseph Cho, Fachrina Dewi Puspitasari, Yu Qiao, Taegoo Kang, Sung-Ho Bae and Choong Seon Hong are with the Kyung Hee University, South Korea (email: chaoningzhang1990@gmail.com; joyousaf@khu.ac.kr; puspitasari-dewi@outlook.com; qiaoyu@khu.ac.kr;  taegoo0205@khu.ac.kr; shbae@khu.ac.kr; cshong@khu.ac.kr.) }
\thanks{Sheng Zheng is with the Beijing Institue of Technology, China (email: zszhx2021@gmail.com.)}
\thanks{Chenghao Li and Chenshuang Zhang are with the KAIST, South Korea (email: lch17692405449@gmail.com; zcs15@kaist.ac.kr.)}
\thanks{Xinru Shan is with the Microsoft STCA, China (email: libchanz@gmail.com.)}
\thanks{Caiyan Qin is with the Harbin Institute of Technology, China (email: qincaiyan@hit.edu.cn.)}
\thanks{Francois Rameau is with The State University of New York - SUNY, South Korea (email: rameau.fr@gmail.com.)}
\thanks{Lik-Hang Lee is with the Hong Kong Polytechnic University, China (email: lik-hang.lee@polyu.edu.hk.)}
}

\newcommand{\pb}[1]{\vspace{0.75ex}\noindent{\bf \em #1}\hspace*{.3em}}
\newcommand{\say}[1]{``#1''}

\markboth{Journal of \LaTeX\ Class Files,~Vol.~14, No.~8, August~2021}%
{Shell \MakeLowercase{\textit{et al.}}: A Sample Article Using IEEEtran.cls for IEEE Journals}


\maketitle

\begin{abstract}
The Segment Anything Model (SAM), developed by Meta AI Research, represents a significant breakthrough in computer vision, offering a robust framework for image and video segmentation. This survey provides a comprehensive exploration of the SAM family, including SAM and SAM 2, highlighting their advancements in granularity and contextual understanding. Our study demonstrates SAM’s versatility across a wide range of applications while identifying areas where improvements are needed, particularly in scenarios requiring high granularity and in the absence of explicit prompts. By mapping the evolution and capabilities of SAM models, we offer insights into their strengths and limitations and suggest future research directions, including domain-specific adaptations and enhanced memory and propagation mechanisms. We believe that this survey comprehensively covers the breadth of SAM’s applications and challenges, setting the stage for ongoing advancements in segmentation technology. 
\end{abstract}



\section{Introduction}\label{introduction}

\begin{figure}[!htb]
     \centering
     \includegraphics[width=\linewidth]{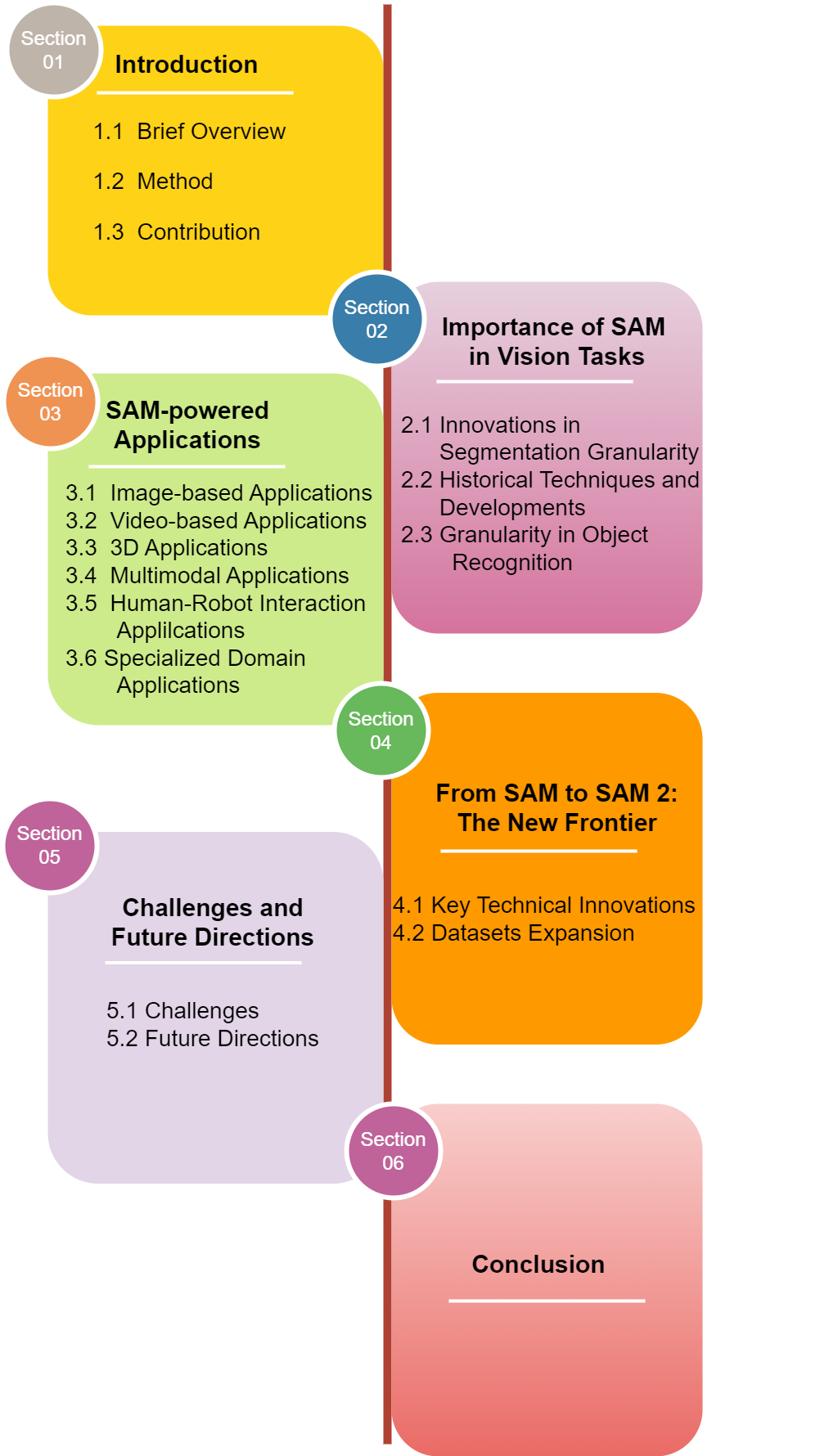}
     
     \caption{Overview of our work}
     \label{fig:vision}
\end{figure}

On July 29, 2024, Meta AI released the second version of the Segment Anything Model (SAM)~\cite{kirillov2023segment}, a major breakthrough in the field of computer vision. Since its inception, SAM has established itself as a pioneering vision foundation model, capable of performing promptable image segmentation tasks that allow the user to interact with the prompts. Specifically, the users can specify the objects of interest with a variety of prompts, such as points, boxes, masks, etc. This innovation has opened new possibilities in image analysis by enabling zero-shot segmentation across a multitude of domains.

Building upon its predecessor, SAM 2~\cite{ravi2024sam} expands these capabilities by incorporating video segmentation and enhancing the granularity of object recognition. SAM 2's innovations include a streaming memory architecture for real-time video processing, improved segmentation accuracy, and a reduced need for user interactions. These advancements address limitations observed in the original SAM, such as handling temporal dimensions in dynamic visual data, and they promise to revolutionize applications in fields like augmented reality, autonomous systems, and medical imaging.

The impact of SAM and its successor, SAM 2, extends far beyond their immediate technical achievements. These models represent a shift towards more interactive and flexible vision systems, capable of understanding and segmenting visual data at unprecedented levels of detail. They empower applications ranging from high-precision medical diagnostics~\cite{zhu2024medical,ma2024segment} to real-time environmental analysis in autonomous vehicles~\cite{li2024fusionsam}. 

Granularity in segmentation plays a vital role in advancing the contextual understanding of vision tasks. By enabling detailed segmentation of objects into their parts, SAM models enhance the capability of AI systems to interpret complex scenes with high fidelity. This granular approach allows SAM and SAM 2 to provide a more nuanced analysis of visual data, enabling the identification of intricate object relationships and interactions within images and videos. Such advancements in segmentation granularity help bridge the gap between machine perception and human-level understanding, which results in breakthroughs in AI across disciplines.

This survey aims to provide a comprehensive overview of the SAM family, focusing on its role in advancing the field of computer vision through the lens of fine-grained and accurate segmentation as a result of combining various prompts. Our work offers an in-depth analysis of both SAM versions, highlighting their contributions to segmentation technology and exploring their applications across multiple domains. By examining the evolution of SAM models, we aim to illustrate their significance in the broader context of AI-driven visual perception and propose future directions for research and development in this area.

\begin{table*}[!htb]
\footnotesize
    \centering
    \caption{Comparison of the extent of discussion between our survey and existing review papers.}
    \label{tab:survey}
    \begin{threeparttable}    
        \begin{tabular}{cp{10.2cm}ccc}  
            \hline
                \multirow{2}*{\bfseries Article}
                & \multirow{2}*{\bfseries Summary of Discussion} & \multicolumn{3}{c}{\bfseries Areas Covered}\\
                \cline{3-5}
                && \textbf{Architecture} & \textbf{Datasets} & \textbf{Applications} \\
            \hline
                 ~\cite{zhang2023comprehensive} & Review of SAM’s architecture and applications across vision tasks & SAM & - & image  \\
                 ~\cite{zhang2024segment} & Survey on SAM's video applications & $\bullet$ & SAM 2 & Video  \\
                 ~\cite{zhou2024image} & Survey on the role of foundation models, including SAM, with promptable segmentation. & - & - & - \\
                 ~\cite{zhang2023towards} & Review of SAM's application in medical image segmentation & SAM & - & Medical \\
                 ~\cite{zhang2024unleashing} & Survey of SAM2’s application in biomedical image and video segmentation & $\bullet$ & - & Biomedical \\
                 Ours & Comprehensive survey on the SAM family (SAM and SAM2), covering advancements, applications, datasets, and future research directions. & $\bullet$ & $\bullet$ & $\bullet$ \\
            \hline
        \end{tabular}
     \begin{tablenotes}
    \small
    \item *\textit{Architecture}: A $\bullet$ covers both SAM and SAM2, while "SAM" refers only to SAM.
    \item **\textit{Datasets}: "SAM2" or specific mentions focus on particular datasets.
    \item ***\textit{Applications}: A $\bullet$ indicates broad applications, while specific mentions (e.g., "Video", "Medical") indicate focused domains.
    \item ****[$\bullet$] and [-] denote available and unavailable discussions, respectively.
\end{tablenotes}

    \end{threeparttable}
\end{table*}

\subsection{Brief Overview}

\pb{Segment Anything Model.} The Segment Anything Model (SAM) is a versatile segmentation model designed to operate across a wide range of visual tasks. By leveraging a promptable architecture, SAM allows users to interactively segment objects using a variety of prompts, providing a level of flexibility and precision that surpasses traditional segmentation methods. This capability is further enhanced in SAM 2, which incorporates video segmentation and a streaming memory architecture to support real-time applications.

\pb{Applications of SAM.}
SAM's ability to perform zero-shot segmentation has made it a valuable tool in numerous fields. Our survey highlights its applications across diverse areas including image-based tasks, video-based tasks, 3D applications, multimodal applications, human-robot interaction, and specialized domain applications. Notably, SAM demonstrates significant potential in medical imaging, where it aids in accurate and efficient analysis of medical scans, and in autonomous systems, where it supports real-time navigation and decision-making. The model's adaptability extends to challenging environments, enhancing its utility across various complex visual contexts.
\subsection{Method}
Our survey utilizes Preferred Reporting Items for Systematic Reviews and Meta-Analyses (PRISMA)~\cite{page2021prisma} framework.
We primarily collect conference and journal papers from well-recognized databases, including IEEE Xplorer, ACM Library, Scopus, and arXiv. The venues of the publications include but are not limited to AAAI, CVPR, ECCV, ICCV, ICLR, NeurIPS, IEEE Robotics and Automation Letters, and IEEE Transactions on Pattern Analysis and Machine Intelligence. 
Note that we include arXiv in our search library list since research in deep learning, particularly the computer vision domain, has developed faster than the peer-reviewed venues can provide.
Given that SAM was only released last year, we restricted the range of publication years of the papers collected from April 2023 to September 2024.
Using the search keywords of \say{segment anything model} and \say{SAM}, we initially curated 428 articles after briefly reviewing the fitness of the publication title.
To ensure that we only review studies closely related to our survey objective, we devise several exclusion criteria, as follows:
\begin{figure}[!htb]
     \centering
     \includegraphics[width=\columnwidth]{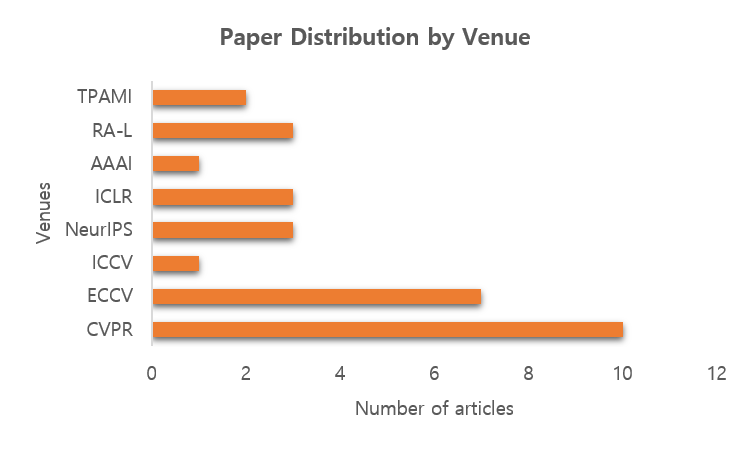}
     \caption{Distribution of selected papers across some key venues}
     \label{fig:vision}
\end{figure}
\begin{itemize}
    \item articles that utilize SAM only as a trivial segmentation tool,
    \item survey and review articles, and
    \item article from arXiv that has not received its first citation despite already being published more than six months ago (to proxy the evaluation towards the quality and usefulness of the papers).
\end{itemize}

\noindent
Implementing this list in the selection process of the content of the abstract and full article, we finally obtained a final list of 192 papers used as the survey's main articles.

\subsection{Contribution}
This survey offers a detailed exploration of the SAM family, focusing on their innovations in segmentation technology and their applications across various domains. We emphasize the significance of SAM's promptable architecture and its role in advancing machine perception through segmentation granularity. Our analysis provides insights into the strengths and limitations of SAM models, offering a foundation for future research and development in AI-driven visual systems. We contribute to the field by exploring the advancements and limitations of SAM models, advocating for potential applications, and suggesting future research directions that address the challenges and opportunities within the scope of segmentation granularity.


\section{Importance of SAM in Vision Tasks}\label{importance}

\subsection{Innovations in Segmentation Granularity}
Despite its popularity as the first foundation model in image segmentation that leverages interactive guidance through prompting, SAM's true capability lies in its granular recognition of objects in the image.
For instance, when given an image of a teapot, SAM can not only segment the whole teapot but also its lid, bowl, and handle simultaneously.
This awareness of granularity emerges as the result of the interaction between ambiguity-aware decoding and semantically unconstrained large-scale masks.

\pb{Ambiguity-aware Decoding.}
The key to making SAM's mask decoder aware of the ambiguous segmentation is parallel decoding.
Parallel decoding has been utilized in many efficient object detection and segmentation models that were built upon transformer~\cite{vaswani2017attention} architecture (e.g., DETR~\cite{carion2020end} and MaskFormer~\cite{cheng2021per}).
The essence of this decoding type is to let the transformer process multiple object queries (analogous to $[class]$ embedding in ViT~\cite{dosovitskiy2020image}) simultaneously, instead of autoregressively.
SAM provides four queries that are specified for the whole class, part class, sub-part class, and additional class (never returned unless multiple prompts are supplied), respectively.
Thus, when decoding, the transformer will calculate the loss between the ground-truth mask and the predicted mask for each of these queries, instead of averaging them into one fixed class.

\pb{Semantically Unconstrained Mask Annotation.}
SAM's ability to recognize a tiny teapot lid on a cluttered kitchen table, for instance, is a result of meticulous manual mask annotation.
One prominent strategy in its annotation is to let all objects in the image be individually segmented without being constrained to the semantic classes they belong to.
Such a technique allows the curation of a huge number of masks because occasionally, there are objects that cannot be sufficiently described by a clear semantic category (or the semantic categorization itself is still unclear).
This semantic relaxation is what sets SAM apart from prior models that start the segmentation task from pre-defined semantic classes contained in the training dataset.
The 4.3M collection of carefully annotated masks is further expanded through semi- and fully-automated annotation stages whose purpose is to increase the mask diversity (particularly in the less apparent objects) and to refine the mask validity in each ambiguous classes, respectively.
The whole mask annotation process is done in a data engine manner which permits SAM to be trained in a fully-supervised manner with an enormous collection of high-quality ground-truth masks.

\subsection{ Historical Techniques and Developments}
Granular segmentation has been explored in several models before SAM.
The attempt to perform this task was first initialized by human parsing task that segments a human body into several limbs.
Segmenting humans to finer granularity was possible with the help of pose estimation.
Pose contains knowledge of human body parts whose keypoint localization aids the segmentation model in recognizing different parts of the body.
JPPNet~\cite{liang2018look} is one example of human parser models that employs two distinct networks, pose and parsing, to learn pixel-wise class and global structure, respectively.
Nevertheless, the dependency on keypoint annotations mask limits the coverage of the pose-parsing method only to naturally-articulated objects (e.g., humans and animals).
Thus, a step further from this technique is to capture objects of various granularity in the image through multi-scale feature extraction.
DeepLab~\cite{chen2017deeplab} is one of the most prominent implementations of object parsing using this method.
The network resizes the input image into various scales that are passed into separate feature extractors.
The resulting granular segmentation is produced by dynamically aggregating these features through the scale attention mechanism.
This multi-scale parsing allows for object-independent granular segmentation.
However, unlike the pose-parsing method, multi-scale learning disregards the information on the structural relationships.
Meanwhile, humans tend to refer to this relationship by nature when they parse an object into parts.
Therefore, following such an intuition, the granular segmentation technique further evolved by incorporating part-object structural information into the segmentation model.
Graphs are one of the common ways to encode this information into the model learning.
For instance, GMNet~\cite{michieli2020gmnet} employs a graph-matching module in the form of adjacency matrices to the aggregated output of the object and part autoencoders.
The loss from this module is specifically used to align the relative spatial relationship between ground truth and predicted parts.
Recently, with the invention of set prediction through DETR~\cite{carion2020end}, granular segmentation responsively adapts the transformer architecture.
For instance, Panoptic-PartFormer~\cite{li2022panoptic} modifies MaskFormer~\cite{cheng2021per} semantic class queries into three granular queries (i.e., thing, stuff, and part).
This technique is perhaps the closest to how SAM implements the whole, part, and sub-part queries in its network.
Despite these advancements, almost all granular segmentation techniques before SAM could not achieve a wide coverage of granularity as they are mainly confined to a small set of training data that is highly skewed towards the granular annotation of humans, animals, and vehicles.

\subsection{Granularity in Object Recognition}
Hobbs'~\cite{hobbs1990granularity} theory of granularity first introduces the decomposition of matter or concept (or granularization) allowing the simplification in comprehending complex phenomena.
This simplification is the foundation for developing many real-world systems through modular executable computations.
Hobbs elaborates that granularity has five characteristics which are constituents, relations, mapping, translation, and idealization.
All these aspects are fundamental ingredients for contextual understanding.
For instance, a tree as a complex ecosystem can be parsed as \say{tree} when a human or computer recognizes the presence of leaves and roots (instances) that are interconnected both spatially and functionally (relation) in almost all kinds of tree (mapping) despite the slight differences on their characteristics in different species of trees (translation and idealization).
Thus, this relational manner of thinking promotes the ability to discern the object on sight.
Note that the existence of four Hobbs' granularity characteristics (exclusive of instances) extends the role of granularity from simply recognizing content to understanding context.

\section{SAM-powered Applications}\label{application}
\begin{figure}[!htb]
     \centering
     \includegraphics[width=\columnwidth]{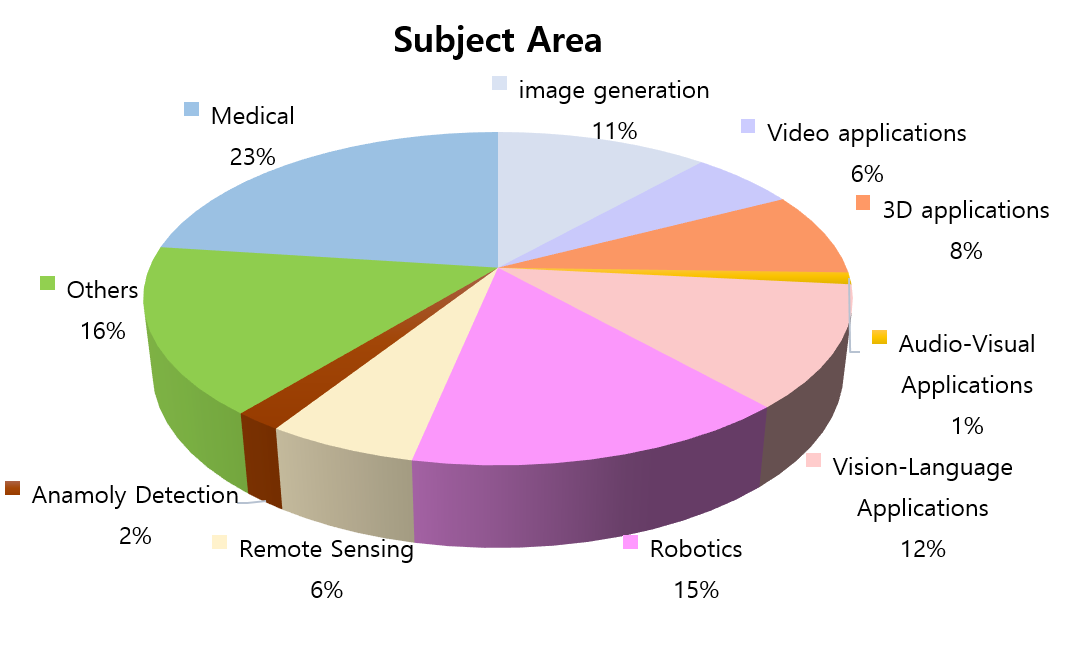}
     \caption{Distribution of studies citing SAM across various application domains, highlighting its versatility and broad impact in fields such as medical imaging, robotics, vision-language applications, and more.}
     \label{fig:vision}
\end{figure}

\begin{figure*}[!htb]
     \centering
     \includegraphics[width=0.9\linewidth]{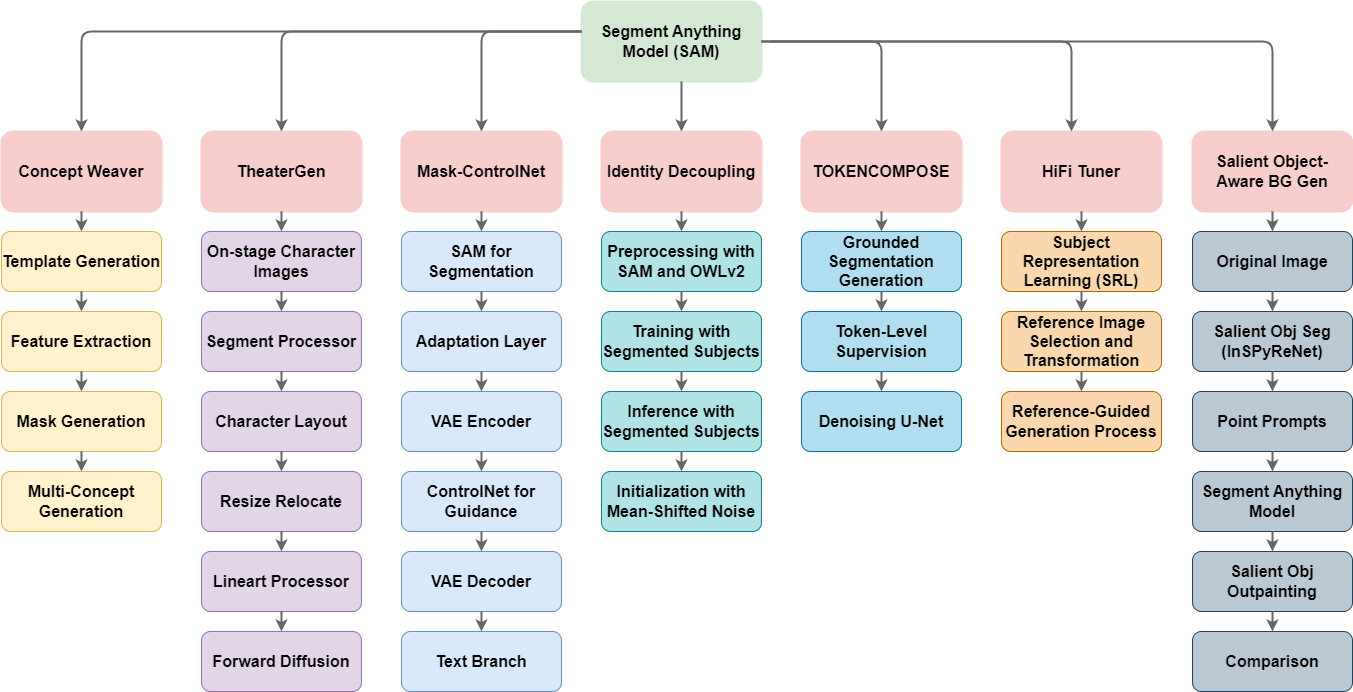}
     \caption{SAM in image generation models. The diagram shows how SAM enhances various image generation tasks by integrating with frameworks like Concept Weaver, HiFi Tuner, and Salient Object-Aware Background Generation. SAM contributes to tasks such as multi-concept generation, character management, object reconstruction, and text-image alignment, emphasizing its versatility in addressing complex vision challenges through precise segmentation.}
     \label{fig:vision}
\end{figure*}

In this section, we survey the literature on the downstream applications of SAM and SAM2. Here, our focus is on their roles across different domains, the purposes they serve, and the benefits they provide.

\subsection{Image-based Applications}

\subsubsection{\textbf{Generation}}

Many works have shown that the granularity provided by SAM greatly improves image-generation applications. For instance, in Concept Weaver\cite{Kwon2024ConceptWE}, SAM generates semantic masks that accurately delineate individual concepts for fusion without blending. Similarly, in TheaterGen\cite{Cheng2024TheaterGenCM}, SAM is used to extract subject masks from character images. These masks are then used to achieve accurate spatial arrangement and segmentation. This contributes to the overall contextual consistency of characters across multiple turns in the generated scene. In another example, HiFi Tuner\cite{Wang2023HiFiTH} uses SAM-generated masks to isolate main objects, guiding loss calculations that ensure high-fidelity representation. Furthermore, in LOOSECONTROL\cite{Bhat2023LooseControlLC}, SAM extracts object segmentation maps and, together with depth conditioning, improves scene generation flexibility. Likewise, in text-guided diffusion models\cite{Eshratifar2024SalientOB}, SAM generates masks for salient objects, preserving boundaries during background generation. In addition, Feng \textit{et al.}\cite{Feng2023RanniTT} in their work utilized SAM to extract object boundaries and attributes, forming control signals that enable more accurate and interactive image editing in diffusion models.

Beyond direct generation, several works leverage SAM for supporting tasks that contribute to improving image quality and composition. For example, in UNIMO-G~\cite{Li2024UNIMOGUI}, SAM is utilized for segmenting image entities. Additionally, it aligns them with textual features in multimodal prompts. This improvement notably increases the overall coherence of the generated output. Similarly, TOKENCOMPOSE~\cite{Wang2023TokenComposeGD} uses SAM's segmentation masks in token-level supervision to improve the composition of multiple visual elements. It ensures better alignment between visual instances and text descriptions.

\subsubsection{\textbf{Inpainting, Stylizing, and Restoration}}
SAM is an important tool for image inpainting since it divides images into coherent patches. These patches can be modified to remove or relocate parts, and the visible regions are then smoothly filled with diffusion models~\cite{yu2023inpaint}. Additionally, to properly execute the inpainting task, it is important to separate foreground from background accurately, which process is known as matting. Different from segmentation, the matting process is more intricate as it has to define an alpha-matte value that defines the opacity of the pixel in the form of trimap. For this reason, generating an alpha-matte mask is more resource-expensive than a normal segmentation mask. Thus, to alleviate this constraint, researchers employ SAM's segmentation power to generate accurate masks that will function as pseudo-trimap~\cite{yao2023matte}. Nevertheless, foreground-background separation can also be achieved through simple approaches, e.g., enhancing the pixel quality of the foreground against the background of a target object choice. In this framework, SAM can be integrated to generate foreground segmentation mask~\cite{yang2023fine}.

\begin{figure*}[!htb]
     \centering
     \includegraphics[width=0.9\linewidth]{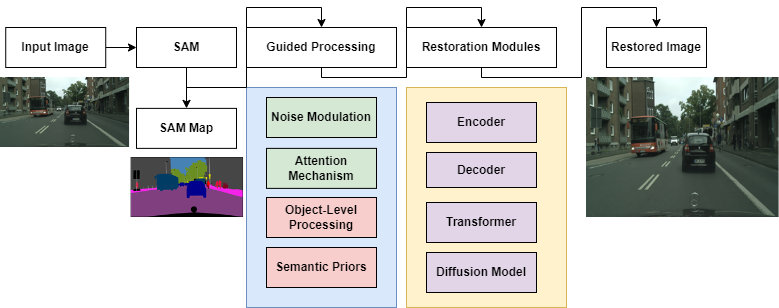}
     \caption{Example of SAM application in image restoration, showing how segmentation maps guide noise modulation and object-level processing for improved restoration quality.}
     \label{fig:vision}
\end{figure*}

For image stylizing, traditionally, applying different styles to individual objects within an image has been an arduous task. This task has now been simplified by SAM’s segmentation capabilities, which enable object-wise stylization and allow precise control over each object. By caching and reusing image encodings~\cite{Ganugula2023MOSAICMS}, SAM speeds up the mask generation process, making it more efficient in handling multiple objects. Moreover, SAM can be directly integrated into stylization workflows, ensuring that each object is accurately stylized independently.

Furthermore, SAM's pixel-to-pixel segmentation effectively avoids color bleeding and preserves image quality in colorization tasks~\cite{Vavilala2023ApplyingAC, Gain2024CCCOC}. Unlike traditional methods that often require manual work, SAM reduces computational costs and delivers visually superior results.

In addition to colorization, SAM has been used in several image restoration works with impressive results. In~\cite{Wang2024SAMDiffSRSD}, SAM creates detailed segmentation masks that guide the noise sampling process. This leads to better detail recovery and higher performance in reducing artifacts. Similarly, in~\cite{Xue2024SegmentationGS}, SAM is used to generate instance segmentation maps. These maps direct the sparse self-attention mechanism to focus on important features. This approach results in better restoration quality and improved visual outcomes. Furthermore, \cite{Jiang2023RestoreAP} employs SAM for interactive and per-object level segmentation. This allows for targeted, high-quality restoration based on the texture and degradation of different objects in the image. Additionally, \cite{Xiao2023ADI} integrates SAM to create high-quality semantic masks that act as priors in a lightweight SAM prior tuning (SPT) component, significantly improving the performance of existing image restoration methods. From these studies, it is clear that SAM is a valuable tool for improving detail recovery, directing attention mechanisms, and providing interactive control for superior visual quality in restored images.

\subsubsection{\textbf{Annotations}}
SAM, due to its granularity, proves to be a game-changer in image and video annotating tasks. It refines pseudo-labels in weakly supervised semantic segmentation by integrating with class activation maps (CAM)~\cite{Chen2023SegmentAM}. This integration improves segmentation accuracy without changing existing methods in image annotation. It generates precise object masks from weak labels like points and bounding boxes. This improves boundary accuracy in segmentation networks~\cite{Jiang2023SegmentAI}. Moreover, in medical imaging, SAM facilitates crowd-sourced annotations by creating dense segmentation masks from sparse inputs. This streamlines the annotation process for 3D deep learning models~\cite{Kulkarni2024AnytimeAA}. SAM also accelerates labeling in multispectral imaging through its integration with MATT~\cite{Gallagher2024AMA}. This approach maintains precision while reducing annotation time. In face anti-spoofing, SAM provides fine-grained pixel-wise masks, boosting model training with detailed annotations~\cite{Chen2023FineGrainedAF}. Furthermore, SAM generates segmentation masks from bounding box prompts, which reduces the cost of labeling data while still keeping performance on par with other methods.

\begin{figure*}[!htb]
    \centering
    \includegraphics[width=0.9\linewidth]{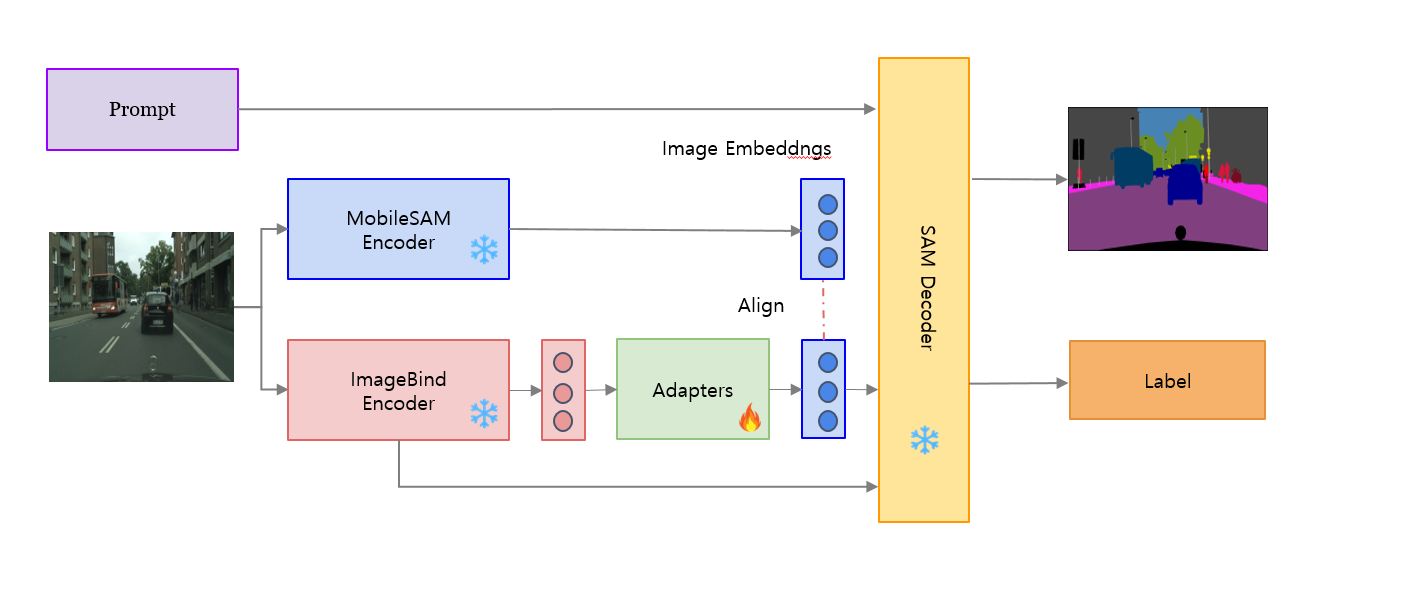} 
    \caption{Example of image and video annotations with SAM assistance.}
    \label{fig:vision}
\end{figure*}
\subsubsection{\textbf{Matching}} 

Among the various composite downstream applications of SAM, image, and video matching are particularly important. In MESA~\cite{zhang2024mesa}, SAM is employed for precise segmentation, improving area-matching accuracy through a multi-relational graph and energy minimization approach. Another study by Zhang \textit{et al.}~\cite{zhang2024mesa} uses SAM for object extraction in unsupervised collaborative metric learning, effectively boosting object embedding and retrieval across scales. In image-to-image matching, SAM generates mask proposals and segments foreground regions, greatly improving region classification accuracy and performance in open-vocabulary semantic segmentation. Building on these capabilities, the MATCHER~\cite{zhang2024mesa} framework integrates SAM for mask proposals from matched points, employing bidirectional matching and robust sampling to improve segmentation quality without additional training. Additionally, SAM serves as a teacher model in~\cite{Wu2023SegmentAM}, guiding local feature learning and improving feature accuracy and robustness for image matching and visual localization. A study on personalized SAM uses a one-shot learning approach for segmentation, accurately identifying and segmenting target objects in diverse contexts~\cite{Zhang2023PersonalizeSA}. 
Lastly, ElSaer \textit{et al.}\cite{ElSaer2024ANF}, present a framework for vehicle inspection, in which SAM refines segmentation masks by eliminating background points, improving image-matching accuracy, and enabling the creation of seamless high-resolution mosaics under challenging lighting conditions.


\subsection{Video-based Applications}
\subsubsection{\textbf{Generation}}

Several papers~\cite{Ma2024MagicMeIV, wang2024customvideo, Wang2024VideoGW, Wu2024DragAnythingMC, Lee2023FastVS, Shen2023StoryGPTVLL, Qin2023DancingAP} have demonstrated the versatility and effectiveness of SAM in various video generation tasks. For example, Ma \textit{et al.}~\cite{Ma2024MagicMeIV}, used SAM to generate segmentation masks from bounding boxes. It isolates subjects from backgrounds for accurate identity customization and face segmentation, which results in improved identity preservation, contextual consistency, and high-resolution outputs. Similarly, Wang \textit{et al.}\cite{wang2024customvideo} employs SAM to segment objects from reference images, which guides the attention mechanism during training. This approach refines subject identity, strengthens the co-occurrence of multiple subjects, and provides better attention control. Furthermore, the paper~\cite{Wang2024VideoGW} uses SAM to segment image frames into foreground and background components. This helps in optimizing the video generation process. It leads to achieving improved frame consistency, optimize video quality, and video generation. In another study, Wu \textit{et al.}~\cite{Wu2024DragAnythingMC}, used SAM to select regions to be controlled in the initial frame and to generate masks for precise motion control. This approach leads to accurate motion control, entity-level manipulation, and improved video quality. The study~\cite{Lee2023FastVS} leverages SAM to refine masks for dynamic regions, which improves object boundary precision and accurately captures moving objects. Consequently, the paper achieves accurate object segmentation, enhanced video realism, and efficient processing. In~\cite{Shen2023StoryGPTVLL}, SAM is employed to generate character segmentation masks, guiding the attention mechanism within the proposed Char-LDM. This results in enhanced character generation, consistent story visualization, and improved reference resolution. Lastly, Qin \textit{et al}\cite{Qin2023DancingAP} used SAM to generate masks for maintaining background consistency and segmenting human figures for accurate inpainting. This achieves better temporal coherence, improved video quality, and accurate human motion representation.

\begin{figure*}[!htb]
     \centering
     \includegraphics[width=1\linewidth]{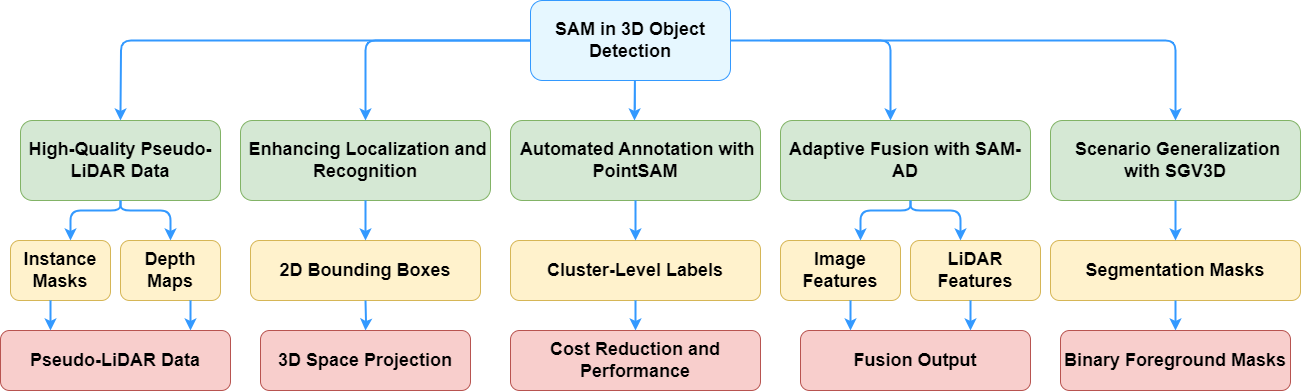}
      \caption{Overview of SAM's role in 3D object detection, showcasing its integration with various methods to refine detection accuracy and performance.}
     \label{fig:vision}
\end{figure*}

\subsubsection{\textbf{Annotation}}

In video annotation, SAM optimizes the EVA-VOS~\cite{Delatolas2023LearningTW} framework by predicting segmentation masks from diverse annotations. This makes labeling videos much faster and more accurate. In a similar vein, SAM improves video text spotting through the SAMText~\cite{He2023ScalableMA} pipeline, producing precise mask annotations. This helps programs to detect and recognize text, resulting in a large-scale dataset with millions of labeled videos.

\subsubsection{\textbf{Tracking}}

Although the original SAM model demonstrated remarkable capabilities in image segmentation, it struggled with temporal consistency in video sequences. Several earlier methods, such as the Track Anything Model (TAM) and XMem~\cite{yang2023track, cheng2022xmem}, were developed to address these limitations. However, the introduction of SAM2 brought significant improvements in video object tracking. For example, in underwater environments~\cite{zhang2024towards} and in Video Object Segmentation (VOS) tasks~\cite{pan2024video, tran20242nd}, SAM2 enhances object segmentation and tracking using its memory-based attention mechanism. It provides better temporal consistency, robustness to occlusions, and improved motion estimation for fast-moving or camouflaged objects. Its zero-shot performance has shown strong results across these tasks.


\subsection{3D Applications}

\subsubsection{\textbf{3D Detection}}

Research shows that SAM helps with 3D object detection. In the paper~\cite{Ding2024VFMM3DRT}, SAM is utilized to generate instance masks that differentiate foreground and background regions. These masks are then combined with depth maps from the Depth Anything Model (DAM)~\cite{yang2024depth} to produce high-quality pseudo-LiDAR data. This approach meaningfully augments 3D detection accuracy and achieves state-of-the-art performance on the KITTI~\cite{geiger2013vision} dataset. This demonstrates its general applicability across various LiDAR-based 3D detectors. Similarly, \cite{Zhang2023FMOV3DFM} employs Grounded-SAM~\cite{GroundedSegmentAnything2023} to generate 2D bounding boxes on images, which are projected into 3D space to strengthen localization and recognition capabilities. This method achieves superior results on datasets like SUN RGB-D~\cite{song2015sun} and ScanNet~\cite{dai2017scannet}. Moreover, \cite{Yang2024MixSupMS} uses PointSAM~\cite{zhou2024point}, derived from SAM, to automate the annotation process by generating coarse cluster-level labels, which leads to significant cost reductions and high performance on benchmarks such as nuScenes~\cite{caesar2020nuscenes} and KITTI~\cite{geiger2013vision}. Furthermore, ~\cite{Song2024RoboFusionTR} devises a customized version of SAM, known as SAM-AD. It extracts robust image features that are fused with LiDAR point cloud features using an adaptive fusion technique, improving robustness and generalization in out-of-distribution noise scenarios. This framework achieves state-of-the-art results on noisy benchmarks like KITTI-C~\cite{kong2023robo3d} and nuScenes-C~\cite{kong2023robo3d}. Finally, SGV3D~\cite{Yang2024TowardsSG} integrates SAM to generate multi-class segmentation masks, for instance, foregrounds. These masks are utilized in the proposed background-suppressed module (BSM) to suppress background features and create binary foreground masks. This method upgrades scenario generalization and achieves superior performance in detecting vehicles, pedestrians, and cyclists in diverse scenes. These studies show that SAM's detailed, context-aware instance masks are key to breakthroughs in 3D object detection. They improve accuracy, reduce the effort needed to label data, and make detection more reliable and adaptable across different situations. This underlines SAM's potential for building complex applications that leverage 3D object detection.

\subsubsection{\textbf{3D (De)composition}}
\begin{figure}[!htb]
     \centering
     \includegraphics[width=1\columnwidth]{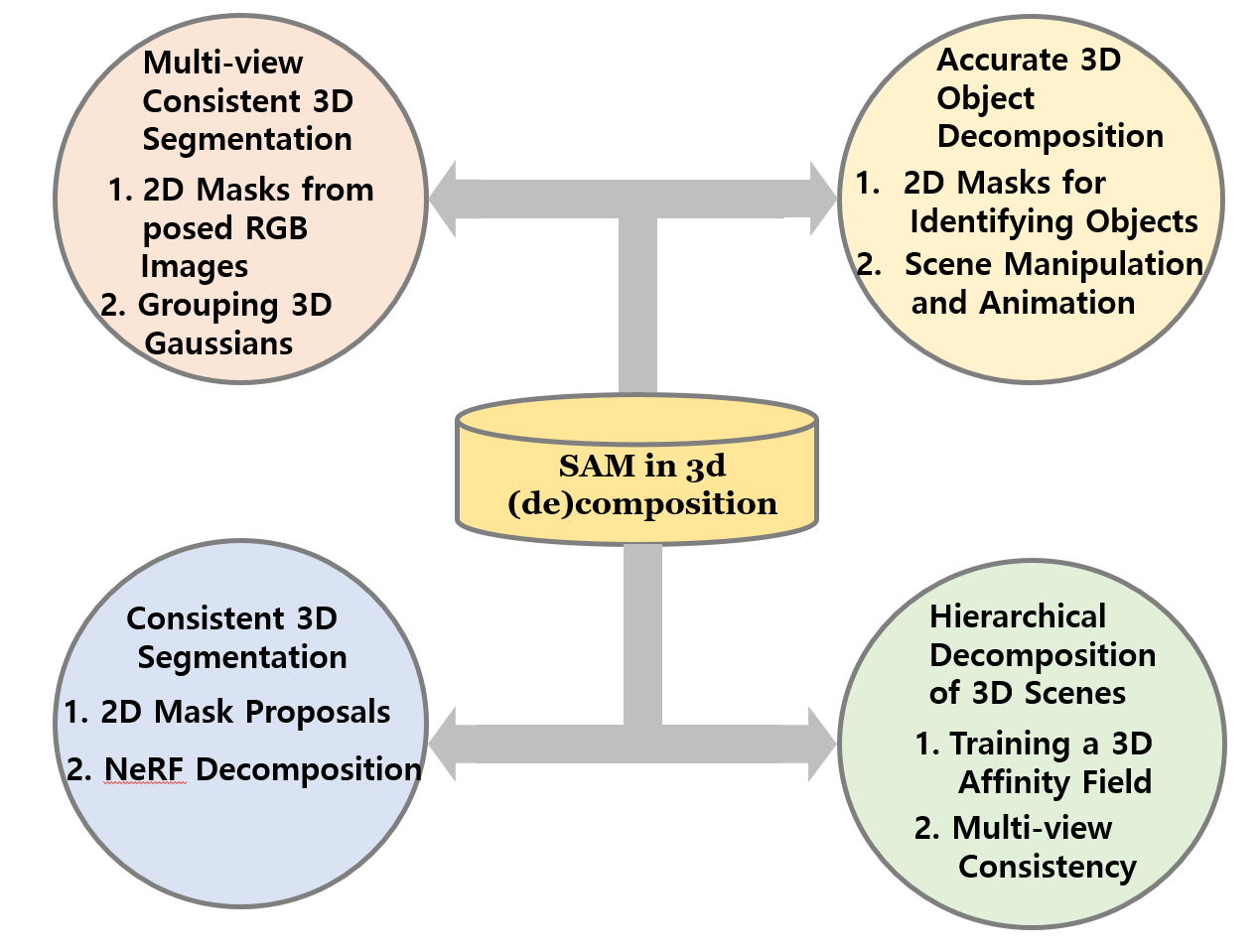}
     \centering
     \caption{SAM in 3D (de)composition}
     \label{fig:vision}
\end{figure}

Recently confronted with numerous challenges in the 3D domain, one of its important areas where SAM assists is 3D (de)composition. First, in the study by  W. Lyu \textit{et al.}~\cite{Lyu2024GagaGA}, SAM is utilized to generate 2D segmentation masks from posed RGB images, in which it serves as a foundation for associating and grouping 3D Gaussians. This leads to multi-view consistent 3D segmentation with high accuracy and consistency, essential for scene manipulation and understanding. Second, X. Lyu \textit{et al.}~\cite{Lyu2024TotalDecomD3} integrates SAM for generating 2D segmentation masks to identify and segment individual objects within a 3D scene, which enables accurate 3D object decomposition and background separation with minimal user input. This approach facilitates detailed scene manipulation, re-texturing, and animation. Similarly, in the work by Zhang \textit{et al.}~\cite{Zhang2023OpenNeRFTO}, SAM is employed to create 2D mask proposals from multiple viewpoints. This aids in the identification and segmentation of objects, thus maintaining consistency and accuracy in 3D segmentation. This method allows the decomposition of neural radiance fields (NeRF) into objects queryable using an open vocabulary, enhancing object manipulation within 3D scenes. Additionally, the research by Kim \textit{et al.}~\cite{Kim2024GARFieldGA} uses SAM to generate 2D segmentation masks. 

These are instrumental in training a 3D affinity field, allowing for the hierarchical decomposition of 3D scenes. This method demonstrates improved multi-view consistency and high-fidelity groupings, which are beneficial for 3D asset extraction and interactive segmentation.


\subsection{Multimodal Applications}

\subsubsection{\textbf{Audio-Visual Applications}}

A growing area of study focuses on utilizing SAM for audio-visual segmentation, as evidenced by recent publications. For instance, in~\cite{Wang2023PromptingSW}, SAM is utilized to refine segmentation tasks by leveraging audio prompts, which sharply improve localization and segmentation in zero-shot and few-shot scenarios. Additionally, in~\cite{Bhosale2024UnsupervisedAS}, SAM is combined with another technique (MoCA), matching audio and visual data to refine pixel-level accuracy without requiring detailed annotations. Similarly, \cite{Liu2023AnnotationfreeAS} adapts SAM with audio embeddings, forming the SAMA-AVS model. It fuses audio-visual features early in the process, and achieves state-of-the-art results. 
\begin{figure}[!htb]
     \centering

     \includegraphics[width=1\columnwidth]{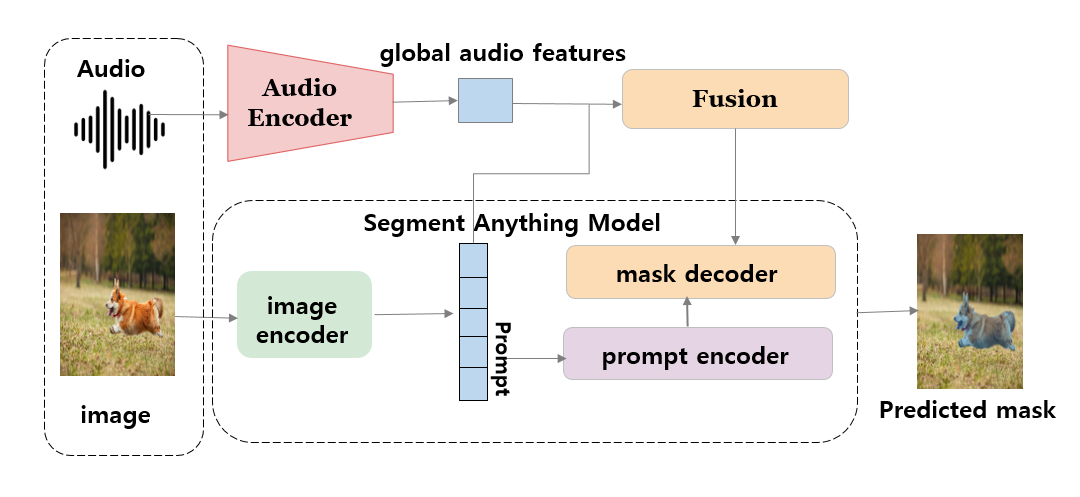} 
     \caption{Example of the integration of SAM in audio-visual application.}
     \label{fig:vision}
\end{figure}

A different approach, \cite{Bhosale2023LeveragingFM}, employs SAM to generate pixel-level masks, effectively segmenting auditory objects without the need for annotated audio-mask pairs, and utilizing audio cues for improved accuracy. Lastly, in~\cite{Mo2023AVSAMSA}, SAM is boosted with pixel-wise audio-visual fusion, significantly improving performance in sound localization and segmentation on datasets like Flickr-SoundNet~\cite{arandjelovic2017look} and AVSBench~\cite{zhou2022audio}.

\subsubsection{\textbf{Vision-Language Applications}}

In vision-language models (VLMs), SAM provides a powerful platform that strengthens their capabilities. Below we have conducted a thorough study of how SAM plays a role in VLM progress. For example, SAM is used to obtain object masks for accurate 3D points extraction. This leads to not only achieving high accuracy in predicting physical properties without labeled data but also outperforming other baseline methods in real-world scenarios~\cite{Zhai2024PhysicalPU}. In a similar vein, in open-world video instance segmentation and captioning, SAM serves as a prompt encoder to generate open-world object queries, resulting in state-of-the-art performance in object detection and captioning tasks~\cite{Choudhuri2024OWVISCapOV}. Additionally, SAM's role in refining hand-object masks noticeably bolsters the localization and captioning of procedural activities in egocentric videos~\cite{Ohkawa2023Exo2EgoDVCDV}. Along with other aspects, SAM's integration with a feature mixer to align regional features with language models facilitates superior regional captioning~\cite{Huang2023SegmentAC}.

Moreover, in the realm of interactive image descriptions, SAM's pixel-level masks for user-defined regions allow for highly detailed image descriptions, thereby improving user interaction~\cite{Wang2023CaptionAI}.

In multi-modal frameworks, Zhou \textit{et al.}~\cite{Zhou2023RegionBLIPAU} and Liu \textit{et al.}~\cite{Liu2024FMFusionIS} discussed how SAM's high-quality object masks lead to improved comprehension, captioning, and superior zero-shot performance in semantic instance segmentation. Additionally, its role as a promptable mask decoder, detailed by Jiao \textit{et al.}~\cite{Jiao2024LumenUV}, boosts vision-centric capabilities across large multimodal models, further supported by its integration in the poly-visual-expert model presented by Fan \textit{et al.}~\cite{Fan2024MouSiPV}.

In tasks involving image fusion, Zhao \textit{et al.}~\cite{Zhao2024ImageFV} introduced how SAM's semantic masks set new accuracy benchmarks, while Zhang \textit{et al.}~\cite{Zhang2024GROUNDHOGGL} and Bao \textit{et al.}~\cite{Bao2024CoReSOT} emphasize its utility in enhancing pixel-level vision-language alignment and reasoning capabilities in segmentation tasks. These improvements are underscored in SAM's application in document OCR, object detection, and visual question answering, where Wei \textit{et al.}~\cite{Wei2024SmallLM} explored its significant contributions.

\begin{figure}[!htb]
     \centering
     \includegraphics[width=1\columnwidth]{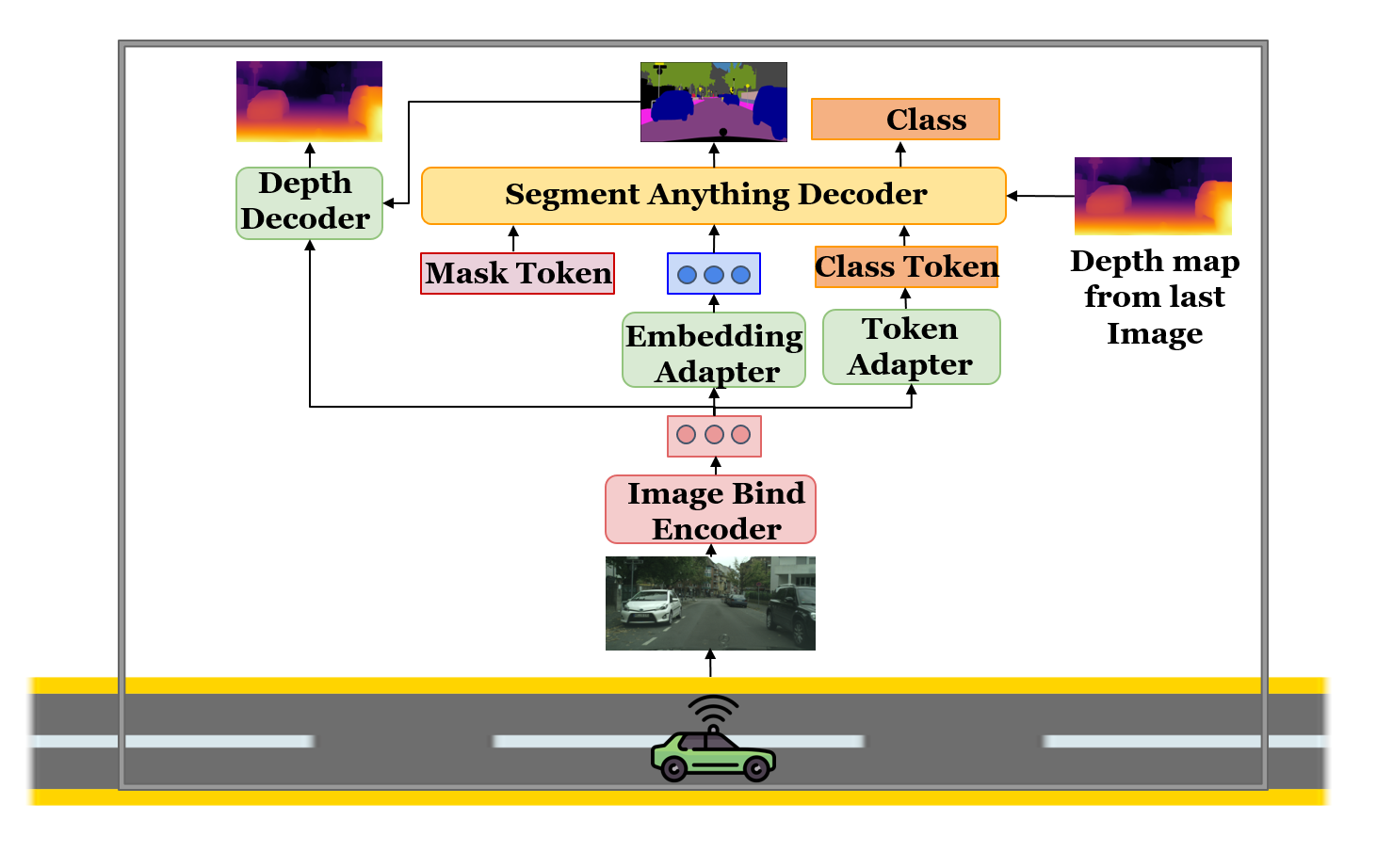}
     \caption{Example of SAM in VLM}
     \label{fig:vision}
\end{figure}
Similarly, many studies~\cite{Lee2024TowardIR, Chen2024SubobjectlevelIT, Kim2024VisionTW} have demonstrated how SAM’s ability to generate semantically meaningful segments accelerates vision-language learning and enhances interpretability. The precision of SAM in refining pseudo-labels for comprehensive scene understanding is thoroughly examined in various works ~\cite{Chen2023TowardsLS}, while its role in producing fine-grained visual knowledge representations for improved multimodal understanding has been explored in recent research ~\cite{Li2024CognitiveVM}.

Lastly, SAM's innovative approach to generating image segments with consistent language semantics significantly improves compositional reasoning and performance across various benchmarks, as explored by Doveh \textit{et al.}~\cite{Doveh2023DenseAA}. This comprehensive review demonstrates SAM's versatility in improving the efficiency, accuracy, and robustness of various vision-language tasks, showcasing its potential to drive superior performance in diverse vision-language domains.


\subsubsection{\textbf{Region-Based Recognition and Generation}}

In \cite{Sun2023AlphaCLIPAC}, Sun \textit{et al.} integrates SAM to generate high-quality pseudo-masks for region-text pairs. This enhances region-based recognition and generation tasks. Similarly, Yang \textit{et al.}\cite{Yang2023MultimodalPM} utilizes SAM for visual perception tasks. This ensures safety and correctness in sequential decision-making tasks by linking visual observations to text-based logic. In another study, Li \textit{et al}~\cite{Li2023MonkeyIR} applies SAM to segment high-resolution images. This contributes to the creation of detailed, multi-level descriptions and improving performance in image captioning and visual question answering. Furthermore, \cite{Xu2023uLLaVAUM} uses SAM to enhance pixel-level understanding. This results in a robust multimodal model with superior fine-grained perception capabilities. \cite{Liu2023LLaVAPlusLT} integrates SAM for interactive segmentation tasks, strengthening the model's toolset and improving performance in human-AI interaction scenarios. \cite{Cai2023LeveragingLL} utilizes SAM to segment images for conversion into SVG format. This facilitates image manipulation based on textual descriptions and optimizing classification and generation capabilities. Lastly, Hsieh \textit{et al.}\cite{Hsieh2023ToolDE} reveals that integrating SAM's segmentation capabilities within a tool documentation framework allows LLMs to perform complex visual tasks effectively without explicit demonstrations.


\subsubsection{\textbf{Pixel-Level Instruction and Fine-Grained Visual Understanding}}

SAM has been pivotal in upgrading various VLMs by leveraging its fine-grained segmentation capabilities. Yuan \textit{et al.}\cite{Yuan2023OspreyPU} used SAM to generate class-agnostic segmentation masks, to allow precise pixel-wise visual instruction tuning. This enhances region-level understanding and fine-grained semantic descriptions. Similarly, Xiao \textit{et al.}\cite{Xiao2023Florence2AA} employed SAM to produce accurate segmentation masks that improve region-specific textual descriptions. Their combined work shows that SAM improves performance in tasks such as object detection, image captioning, visual grounding, and segmentation.
Additionally, Liu \textit{et al.}~\cite{Liu2023InternGPTSV} demonstrated that the integration of SAM enables precise selection and manipulation of visual content based on pointing instructions. This improves both the efficiency and accuracy of tasks like object removal and image editing.
\subsubsection{\textbf{Refining Vision Vocabulary and Visual Grounding}}

Xu \textit{et al.}\cite{Xu2023PixelAL} and Rasheed \textit{et al.}\cite{Rasheed2023GLaMMPG} both utilize SAM for pixel-wise localization and segmentation. It generate location prompts that align words with specific pixel locations and integrating segmentation masks into natural language responses. This upgrade tasks like referring localization, dense object captioning, visually grounded conversations, and image captioning. Similarly, Wei \textit{et al.}\cite{Wei2023VarySU} and Zhao \textit{et al.}\cite{Zhao2023BuboGPTEV} employ SAM to process high-resolution inputs and extract detailed semantic masks. This enables the merging of new vision vocabularies with existing ones and improves visual understanding in tasks such as document OCR, chart understanding, and multi-modal interaction. In another study, Chen \textit{et al.}\cite{Chen2023MitigatingHI} used SAM to guide LVLMs. This approach significantly reduced hallucinations and improved contextual accuracy. Lastly, Li \textit{et al.}\cite{Li2023EmpoweringVM} integrates SAM within the Visual Prompt Generator Complete module to recover missing visual details. Their study shows the performance was increased in tasks requiring detailed visual reasoning and contextual understanding.


\begin{figure*}[!htb]
     \centering
     \includegraphics[width=1\linewidth]{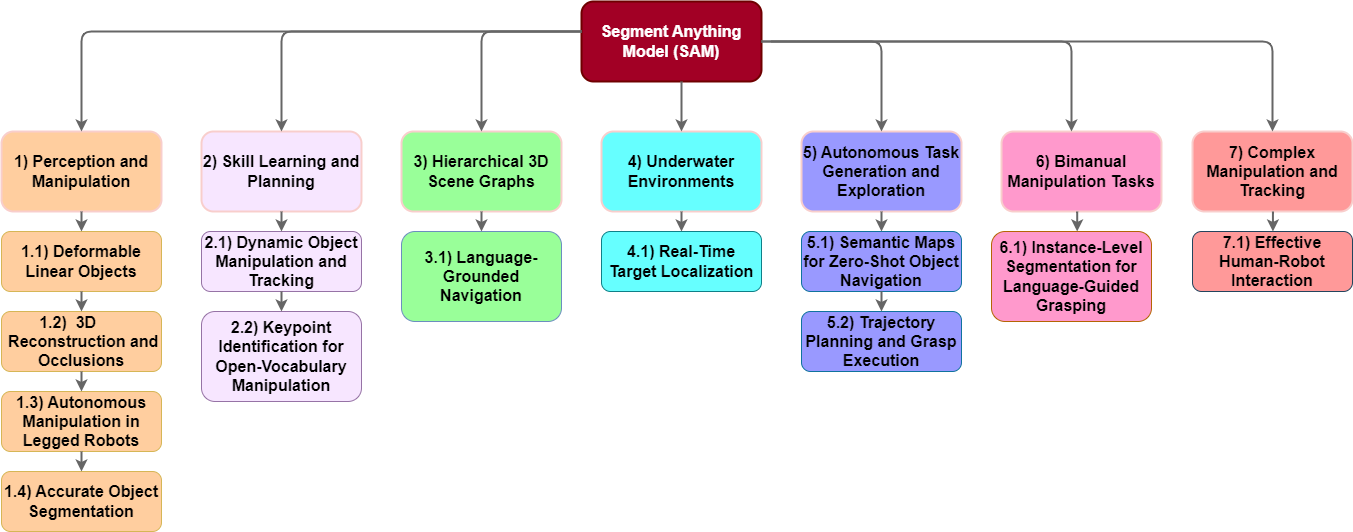}
      \caption{Illustration of SAM applications in human-robot interactions, highlighting its role in various tasks.}
     \label{fig:vision}
\end{figure*}

\subsection{Human-Robot Interaction Applications}

In robotics, several works~\cite{Sun2024ARD, Liu2024VisualWC, Kumar2024PracticeMP, Xiang2024GraspingTO} have explored the integration of the Segment Anything Model (SAM) to refine various applications through its precise segmentation capabilities. For instance, Sun \textit{et al.}~\cite{Sun2024ARD}, employed SAM to improve the perception and manipulation of deformable linear objects. This enables robust 3D reconstruction and addressing challenges like occlusions. In another application, SAM has also been used to improve autonomous manipulation in legged robots, providing accurate object segmentation and facilitating interaction with diverse objects~\cite{Liu2024VisualWC, Kumar2024PracticeMP, Xiang2024GraspingTO}.

In skill learning, SAM helps robots with planning and manipulation tasks. It optimizes their ability to learn on their own~\cite{Li2024GrowingFE, Liang2024LearningTL}. Its integration into mobile platforms expands the potential for dynamic object manipulation and tracking~\cite{Zhang2024InnovativeIO, Gao2024BiKVILKV, Xu2024TowardsUI}. Additionally, SAM aids in keypoint identification for open-vocabulary manipulation, which enhances motion planning~\cite{Liu2024MOKAOR, Zheng2024GaussianGrasper3L, Lin2024TwistingLO}. In a similar vein, hierarchical 3D scene graphs also benefit from SAM's segmentation. It enables more effective language-grounded navigation in complex environments~\cite{Werby2024HierarchicalO3, Tie2024O2VMappingOO, Huang2024IVLMapIV}.

Also, some studies~\cite{Wu2024VoroNavVZ, Yang2024BBSEAAE, Kuang2024OpenFMNavTO, Zhang2023InteractiveNI} have shown that SAM assists in autonomous task generation and exploration, which leads to better adaptability in robotic systems. It is also crucial in generating semantic maps for zero-shot object navigation, which leverages topological and semantic information~\cite{Kuang2024OpenFMNavTO, An2023RGBManipMI}.


In grasping tasks, SAM optimizes trajectory planning by precisely segmenting objects in 3D environments and in open-world scenarios~\cite{Lin2023GestureInformedRA}. This improves grasp execution and collision avoidance~\cite{Tang2023GraspGPTLS, Barad2023GraspLDMG6}. 
Beyond the aforementioned applications, SAM has also been used in some other complex manipulation and tracking tasks, such as folding cloths~\cite{Hato2024AutonomousMO}, managing articulated objects~\cite{Buchanan2023OnlineEO}, and monitoring laboratory environments~\cite{Hato2024AutonomousMO}.
Furthermore, in the autonomous driving scenario, Zhou \textit{et al.}\cite{zhou2023dsecmos} curates an event-based segmentation dataset from object detection results in stereo event camera data by utilizing SAM. Additionally, SAM is also helpful in enhancing the calibration performance between LiDAR and the camera. Luo \textit{et al.}\cite{luo2023calib} demonstrate this in~\textit{Calib-Anything}, where they leverage SAM's generalizability to perform semantic segmentation and generate masks. These masks are then used as a container to perform point cloud optimization and generate consistency scores.


\subsection{Specialized Domain Applications}

\subsubsection{\textbf{Remote Sensing}}

SAM is a popular choice in remote sensing applications because it provides detailed and comprehensive contextual information. Ren \textit{et al.}\cite{ren2023segment} demonstrated that SAM can perform well in segmenting minuscule objects captured by the overhead camera. Meanwhile, they also discovered that SAM failed to segment slender target objects, such as road and farm parcel boundaries. 
In a related study, \textit{Yu et al.}\cite{yu2023sea} experiments with SAM and CLIP to segment sea ice and show that such a combination can distinguish ice types. Based on this potential, researchers have extended SAM's applications for overhead imagery. For example, Wang \textit{et al.}\cite{wang2023scaling} utilizes SAM to generate a large-scale remote sensing segmentation dataset that surpasses prior datasets in terms of size and information such as categories and locations. 

SAM has been instrumental in improving the accuracy and performance of semantic segmentation in remote-sensing images. For example, the MeSAM~\cite{Zhou2024MeSAMME} model adjusts SAM to process optical remote sensing images more efficiently by keeping high-frequency features intact and using multiscale convolutional kernels. In a similar vein, RSAM-Seg~\cite{Zhang2024RSAMSegAS} improves SAM by adding domain-specific knowledge and scaling modules, making it outperform the original SAM and other models in tasks like cloud, building, field, and road segmentation. Additionally, there is also \textit{Text2Seg}, which was proposed by~\cite{zhang2023text2seg}. This pipeline takes advantage of SAM's zero-shot segmentation capabilities, it explores the applicability of multiple visual foundation models to facilitate text-guided semantic segmentation tasks in remote sensing images with minimum model adjustments~\cite{Zhang2023Text2SegRS}.

Change detection in remote sensing images is another area where SAM excels. For instance, in zero-shot change detection, SAM has been used to create the AnyChange~\cite{Zheng2024SegmentAC} model, which detects changes without needing extra training and sets new performance benchmarks. Moreover, integrating SAM into the proposed Time Travelling Pixels (TTP) method for bi-temporal change detection has also enhanced both the accuracy and efficiency of spotting significant changes over time~\cite{Chen2023TimeTP}. Additionally, the Segment Change Model (SCM)~\cite{Tan2023SegmentCM} takes advantage of SAM's detailed segmentation to greatly improve performance in unsupervised change detection.

Beyond the aforementioned applications, SAM continues to demonstrate versatility across other remote sensing tasks. In cropland mapping, using SAM with prompt learning methods improves both accuracy and efficiency~\cite{Tao2023UsingGL}. Similarly, fine-tuning SAM for river water segmentation not only improves performance but also helps in developing new datasets~\cite{Moghimi2024ACP}. For thermal imagery taken by aerial field robots, SAM ensures more precise and consistent labels~\cite{Lee2024SemanticsFS}. It works well with vision-language models for tasks like zero-shot image classification, retrieval, segmentation, and visual question answering in satellite images~\cite{Mall2023RemoteSV}. 
\begin{figure}[!htb]
     \centering
     \includegraphics[width=1\columnwidth]{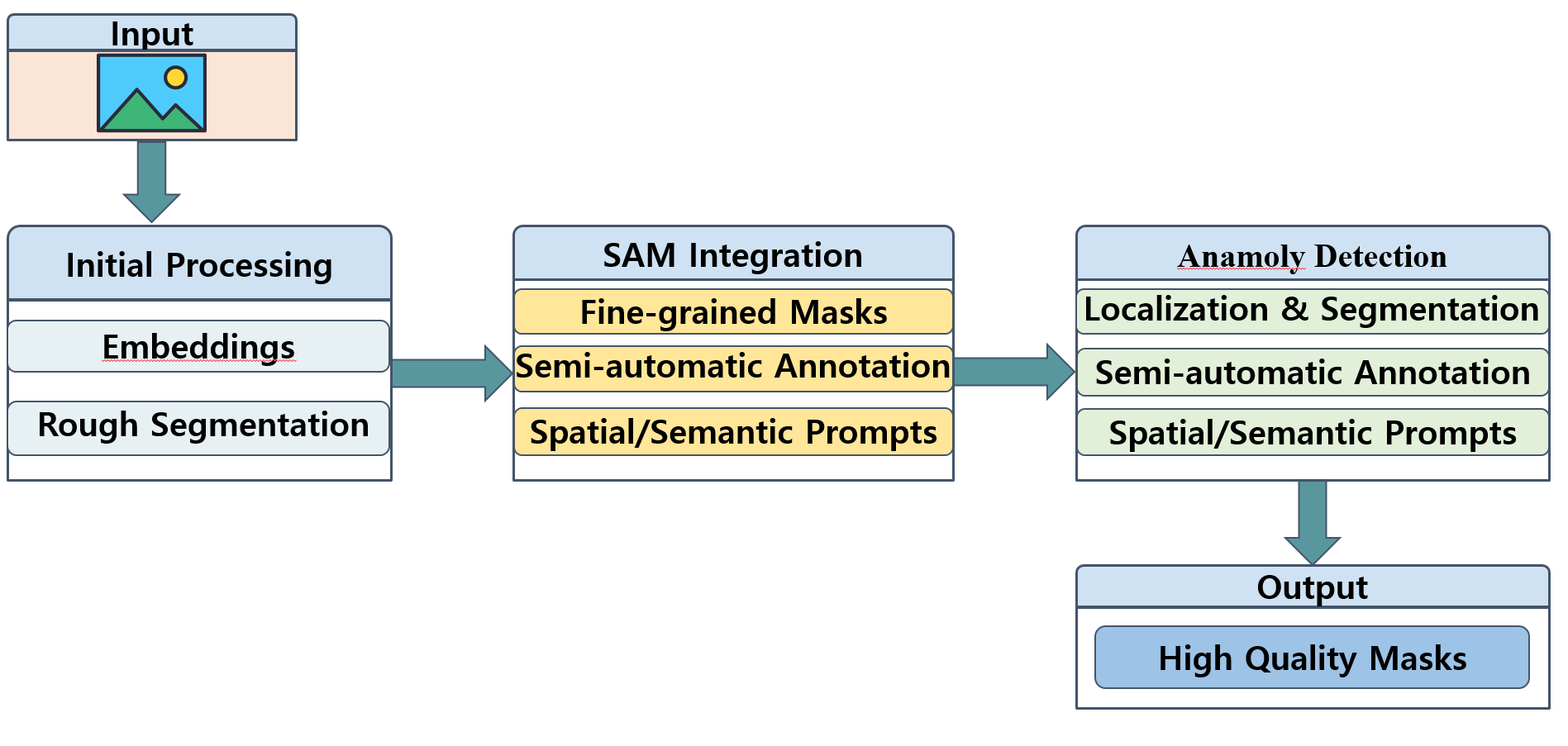}
     \caption{SAM in anamoly detection}
     \label{fig:vision}
\end{figure}
These diverse uses show how valuable SAM is for improving segmentation accuracy, efficiency, and performance in various remote sensing applications.

\subsubsection{\textbf{Anamoly Detection}}

Anomaly detection is one of the many areas where SAM has been effectively utilized. Li \textit{et al.}~\cite{Li2024ClipSAMCA} used SAM to refine rough segmentation results provided by CLIP, producing precise masks guided by spatial prompts. This refinement elevates anomaly localization and segmentation performance on datasets such as MVTec-AD~\cite{bergmann2019mvtec} and VisA~\cite{zou2022spot}. Subsequently, GPT-4V-AD~\cite{Zhang2023ExploringGP} draws upon SAM to generate semantic-level region divisions, enhancing the model's focus on relevant areas for detecting anomalies. Furthermore, Cao \textit{et al.}~\cite{Cao2023SegmentAA} proposed a SAA+ framework that incorporates SAM to refine bounding-box-level anomaly region candidates into high-quality pixel-wise segmentation masks using hybrid prompts, which reduces false alarms and improves anomaly localization accuracy on benchmark datasets.

Another work called DefectSAM~\cite{Hu2023SegmentAI} leverages SAM to segment defects from infrared thermal images using expert-provided prompts. This leads to superior defect detection rates and more accurate defect size estimations. Similarly, in a two-stream lightweight model for anomaly detection~\cite{Li2024AST}, SAM acts as a teacher network. It guides the training process and generates high-quality segmentation masks that guarantee efficient and accurate anomaly detection.
Upgrade in the WinCLIP framework~\cite{Baugh2023ZeroShotAD} is observed by incorporating SAM for foreground extraction and improved localization of smaller and subtler anomalies, leading to higher F1-max scores on the VisA~\cite{zou2022spot} dataset. Finally, in a multi-scale memory comparison framework for zero-/few-shot anomaly detection~\cite{Huang2023MultiScaleMC}, SAM is key to segmenting complex scenes into individual objects. It facilitates the framework to focus on specific anomalies within each object and improve overall detection and localization accuracy.


\subsubsection{\textbf{Farming}}
Agriculture is also a potential scenario for SAM application. \cite{yang2023sam} examines the application of SAM in precision livestock farming (PLF), particularly chicken. They found that the high IoU value of segmentation by SAM compared to other methods contributes to the success of following tracking tasks. Nevertheless, due to its generality, SAM still fails to solve the common issues in PLF, e.g., occlusion due to the dense environment and behavior variance. 
Additionally, \cite{williams2023leaf} experimented with post-processing, e.g., color checking, shape correction, and mask filtering, to relax the SAM's class-agnostic assumption for the leaves classification task. 
Further, \cite{jain2023vision} ideate the integration of SAM for the packed meal packaging industry, particularly to assist the conceptualization of target food composition.


\subsubsection{\textbf{Others}}
In a space exploration scenario, \cite{julka2023knowledge} attach an additional decoder into SAM's architecture, whose purpose is to learn domain-specific information. With a slightly different approach, \cite{giannakis2023deep} experimented to segment the crater with SAM, which output mask is passed through post-processing steps, \textit{i.e.}, circular shape filtration, boundary extraction, and ellipse fitting. 

SAM also offers promising results in the semantic communication domain compared to traditional improvement approaches that are spectrum-limited and need high power consumption. Capitalizing on SAM's generalizability and promptability, Tariq \textit{et al.}\cite{tariq2023segment} leverage SAM in their semantic communication framework to transmit only selected information at high broadcasting quality. 

SAM applicability also extends to civil construction scenarios. Ahmadi \textit{et al.}\cite{ahmadi2023application} introduces SAM to segment structural defects, \textit{e.g.}, cracks. Their study found that SAM performed better than UNet in detecting longitudinal cracks because SAM's detection is more spatially oriented.


\section{From SAM to SAM 2: The New Frontier}\label{sam2_advancements}
\begin{figure*}[!htb]
     \centering
     \includegraphics[width=0.9\linewidth]{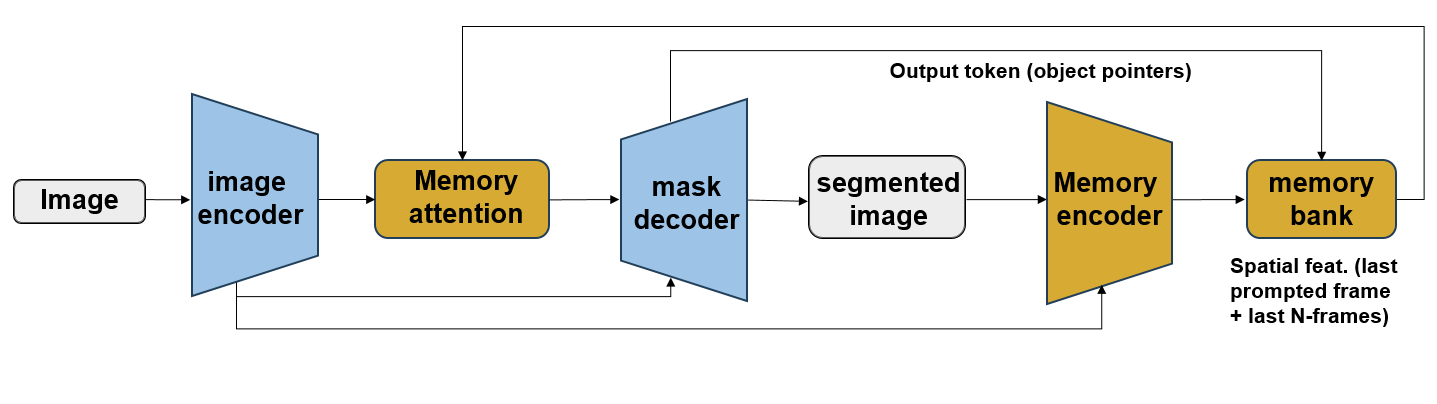}
     \caption{Schematic representation of the SAM 2 architecture illustrating the process flow from image input through memory encoding to generate segmented outputs.}
     \label{fig:vision}
\end{figure*}

The Segment Anything Model (SAM)~\cite{kirillov2023segment} has redefined the segmentation field by introducing a prompt-based method that is highly efficient and can achieve zero-shot performance in a range of tasks. Its successor, SAM 2~\cite{ravi2024sam} has expanded its abilities greatly, especially in dealing with video data and ensuring consistency over time. SAM 2 has a memory feature that enables it to not just isolate single frames, but also to follow and isolate objects throughout video sequences. This progress tackles important obstacles in both static and dynamic segmentation situations, establishing SAM 2 as a significant contributor in areas like self-driving cars~\cite{li2024fusionsam}, medical imaging~\cite{yu2024sam,shen2024performance}, and real-time video analysis~\cite{wang2024robust}. Significantly, SAM 2 has demonstrated enhanced zero-shot performance in challenging environments such as surgical videos in medical and robotic surgery~\cite{shen2024performance}, outperforming current models in accuracy and robustness.

SAM 2's ability to sustain object tracking~\cite{wang2024articulated} through memory attention mechanisms allows it to manage lengthy video sequences while maintaining segmentation quality, even when faced with occlusion and motion blur. The progression from SAM to SAM 2 not only brings in new advanced features but also moves towards being more generalizable and adaptable across various domains, particularly where fine-tuning is limited or infeasible. In this section, we examine the major technological shifts and dataset expansions that define SAM 2's evolution, which set a new standard for video segmentation technologies.

\subsection{Key Technical Innovations}
SAM 2 brings in numerous architectural and technical advancements that greatly improve its capabilities compared to SAM. In this subsection, we present the core innovations that contribute to SAM 2's superior segmentation performance across diverse tasks and domains.

\subsubsection{\textbf{Unified Segmentation Framework}}
SAM was introduced as a foundational model for segmentation that is capable of performing promptable segmentation tasks across different domains. Its architecture is based on a ViT backbone, which processes images and generates segmentation masks using interactive prompts. SAM 2 extends this concept by introducing a unified segmentation framework that handles both images and videos. Leveraging the Promptable Visual Segmentation (PVS) approach, SAM 2 efficiently integrates spatial and temporal data, which allow it to seamlessly transition between static images and dynamic video sequences, making it suitable for a broader range of applications.

\subsubsection{\textbf{Enhanced Prompting Mechanisms}}
SAM 2 retains the core functionality of its predecessor by allowing users to provide prompts such as points, boxes, and masks. However, it significantly improves upon SAM's original prompting mechanisms by incorporating a memory attention system, which ensures that segmentation remains consistent across frames in a video. In scenarios where objects may disappear from the frame or become occluded, SAM 2 can use previously stored information from earlier frames to maintain segmentation accuracy. This enhancement is particularly useful in complex scenes, as it allows for refining the segmentation with additional input in later frames, which ensure robust and contextually accurate segmentation throughout video sequences.

\subsubsection{\textbf{Streaming Memory Architecture}}
One of SAM 2's key innovation is that it introduces a novel streaming memory architecture that efficiently manages temporal information in video data. Unlike SAM, which processes each frame independently, SAM 2 segments the initial frame using the input prompt and leverages this result to guide the segmentation of subsequent frames. This architecture enables SAM 2 to maintain coherence and temporal consistency across video frames, making it highly effective in handling dynamic environments with moving objects and scene changes. By storing keyframe information in a memory bank, SAM 2 can refine segmentation over time, ensuring that previous segmentations contribute to the accuracy of later frames, even in the presence of occlusions or object transformations.

Additionally, this architecture optimizes computational efficiency by reducing the need to process every frame from scratch, allowing SAM 2 to scale to larger datasets and handle longer video sequences with greater speed and accuracy. This makes it particularly suitable for large-scale applications such as autonomous driving, surveillance, and medical video analysis, where real-time performance and long-term consistency are critical.

\subsection{Dataset Expansion}
Both SAM and SAM 2 have been trained on extensive datasets, but SAM 2's training incorporates the new SA-V (Segment Anything - Video) dataset, which includes a diverse array of video data with pixel-level annotations. This dataset enhances SAM 2's ability to generalize across domains and improve performance in challenging segmentation tasks, such as those involving occlusions and motion blur. We summarize the difference between SAM's and SAM 2's datasets as follows.

\subsubsection{\textbf{SAM: SA-1B Dataset}}

The SA-1B dataset is a cornerstone of the original SAM. It was designed to provide a comprehensive foundation for developing a model capable of zero-shot segmentation across a diverse array of scenarios. The key characteristics of the SA-1B dataset are as follows.

\begin{itemize}
\item \textbf{Scale and Diversity.} The SA-1B dataset contains over 1 billion masks, covering a wide spectrum of object categories and scene types. This immense scale allows SAM to generalize effectively to unseen data and adapt to various segmentation tasks.

\item \textbf{High Resolution.} Images in SA-1B are high-resolution, enabling the model to learn fine-grained details crucial for accurate segmentation.

\item \textbf{Automatic Mask Generation.} The dataset utilizes automatic mask generation techniques to ensure broad coverage of objects, minimizing the need for extensive manual annotation while maintaining high-quality training data.

\item \textbf{Geographical Representation.} SA-1B encompasses data from a variety of geographical locations, ensuring that SAM can handle diverse visual contexts and reducing biases associated with location-specific features.

\end{itemize}

\subsubsection{\textbf{SAM 2: SA-V Dataset}}

The SA-V dataset is introduced with SAM 2 to extend the model's capabilities to video segmentation, addressing the need for temporal understanding in dynamic scenes. SA-V provides 53 times more annotated masks than any previous video segmentation dataset, such as YouTube-VOS and DAVIS. 
\begin{figure}[!htb]
     \centering
     \includegraphics[width=1\columnwidth]{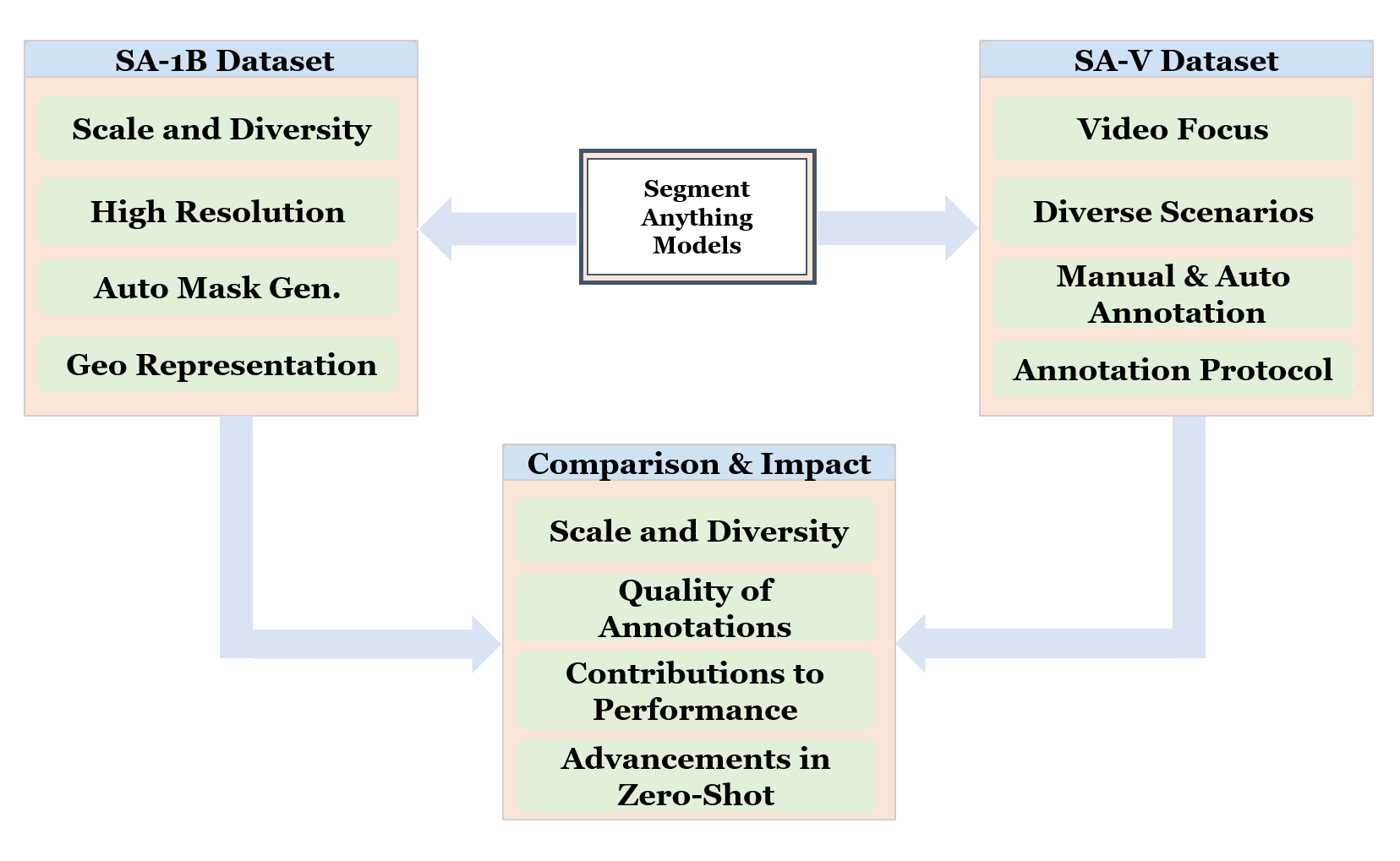}
     \caption{Datasets used for training SAM and SAM 2}
     \label{fig:vision}
\end{figure}
The key characteristics of the SA-V dataset are as follows.

\begin{itemize}
\item \textbf{Video Focus.} Unlike SA-1B, SA-V is centered around video data, enabling SAM 2 to tackle the complexities of motion and temporal consistency in segmentation tasks.

\item \textbf{Diverse Scenarios.} The dataset includes videos from various environments, such as indoor, outdoor, and complex dynamic scenes, providing a robust platform for training the model to handle real-world challenges.

\item \textbf{Manual and Automatic Annotations.} SA-V combines both manually annotated and automatically generated masklets, ensuring high-quality segmentation data while also leveraging the efficiency of automatic processes.

\item \textbf{Annotation Protocol.} The annotation process for SA-V is multi-phased, with an initial focus on frame-by-frame manual annotations followed by the integration of automated tools like SAM 2 to propagate masks across frames. This iterative approach enhances annotation speed and accuracy.

\end{itemize}

\section{Challenges and Future Direction}\label{challenges_and_limitations}

In this section, we present various open research challenges faced by SAM and SAM 2 in real-world applications, particularly in complex and domain-specific tasks. Additionally, we present potential future research directions that aim to address these limitations and improve the models' overall performance and adaptability.
\subsection{Challenges}

As the title suggests, SAM~\cite{kirillov2023segment} is claimed to segment anything in images. However, it remains unclear whether SAM can work as a versatile segmentation model in real-world scenarios. Therefore, numerous works have been conducted recently to evaluate its performance in various scenarios, including medical images, camouflaged and glass objects, etc. With the introduction of SAM 2~\cite{ravi2024sam}, while further studies have shown enhanced capabilities in both image and video segmentation, they also highlight significant challenges such as struggles with unclear boundaries and limited prompt support, particularly in complex medical and camouflaged object scenarios.

\subsubsection{\textbf{Medical Imaging}}

Despite its crisp segmentation results in most common scenarios, raw segmentation outputs of SAM in the medical domain are found to be inferior in accuracy (Dice score) to fully-supervised models, e.g., U-Net~\cite{he2023accuracy}. Moreover, different organs produce different score discrepancies. For instance, the two largest gaps (70\%) of accuracy between SAM and U-Net are observed in the segmentation for the pancreas~\cite{he2023accuracy} and liver~\cite{hu2023sam}, while the two smallest gaps (30\%) are found from polyps~\cite{zhou2023can} and lung nodules~\cite{he2023accuracy}. This underperformance is potentially caused by the different nature of the objects being segmented. Objects with more apparent boundaries produce higher segmentation accuracy than those with obscure boundaries~\cite{huang2023segment}. For example, SAM can segment benign tumors more easily than malignant tumors because the latter do not have clear boundaries due to metastasis to the neighboring tissue~\cite{hu2023breastsam}. Nevertheless, despite its limitations compared to fully-supervised models, SAM segmentation still outperforms non-deep learning segmentation methods, such as FSL BET~\cite{mohapatra2023brain}.

SAM 2 addresses some these challenges by providing more robust segmentation options, especially in 3D settings\cite{yu2024novel}. Nonetheless, studies ~\cite{ma2024segment, chen2024sam2} indicate that in certain cases, particularly with lower contrast modalities like CT and ultrasound, SAM 2 may perform worse than its predecessor. It does, however, show comparable or superior performance in MRI settings. Despite the advancements, both SAM and SAM 2 continue to struggle with over-segmentation issues, particularly when dealing with organs that have unclear boundaries.

\subsubsection{\textbf{Challenging Objects and Scenes}}

Beyond medical images, many real-world segmentation tasks are exposed to challenging conditions that weaken the segmentation capability of SAM. Minuscule and slender objects~\cite{ji2023sam, ji2023segment, ren2023segment}, objects with obscure boundaries~\cite{jie2023sam, ji2023sam, ji2023segment}, occluded objects in dense environments~\cite{yang2023sam}, camouflaged objects~\cite{tang2023can, ji2023sam, ji2023segment}, and transparent objects~\cite{han2023segment, ji2023segment} are a few examples of scenarios where SAM segmentation outputs are rather inaccurate. For this reason, the raw output map of SAM cannot be used directly in object counting tasks~\cite{ma2023sam}.

Researchers have evaluated that SAM struggles to achieve satisfying performance in challenging object detection tasks. For instance, \cite{tang2023can} found that SAM underperformed compared to other state-of-the-art object detectors in camouflaged object detection tasks. A similar result was also discovered by~\cite{han2023segment} in transparent object detection tasks. \cite{ji2023sam, ji2023segment} also observed SAM's unsatisfactory performance in unprompted challenging object detection tasks in various scenes, e.g., camouflaged animals, industrial defects, and medical lesions. Additionally, SAM segmentation results also underperformed in shadow detection as observed by~\cite{jie2023sam}. \cite{ma2023sam} also discovered that the raw mask output of SAM still performs poorly in the object counting task, specifically on tiny and dense objects.

SAM also becomes part of a segmentation framework by~\cite{guo2023prompt} for rainy scenes. Here, using entropy filtering, they pre-processed the image to locate the correct anchor prior to feeding into SAM to refine SAM's mask outputs.

\subsubsection{\textbf{Prompt Dependence}}

SAM's performance is dependent on the types and magnitude of prompts fed to the prompt encoder. In general, automatic prompting yields unsatisfactory segmentation results~\cite{huang2023segment}. Meanwhile, box prompts produce higher segmentation accuracy compared to point prompts~\cite{cheng2023sam, wang2023sam, wald2023sammd, mattjie2023zeroshot}. However, accuracy from point prompts can be increased by applying more points on the target object~\cite{cheng2023sam, hu2023sam, mazurowski2023segment, wald2023sammd, mattjie2023zeroshot, huang2023segment}. Additionally, box prompts and point prompts can be combined to produce better accuracy, but not when applied simultaneously~\cite{huang2023segment}. Improvement is apparent when the box prompt is applied in the initial prompting stage, while the point prompt is administered during the mask refinement stage~\cite{mattjie2023zeroshot}.

\subsubsection{\textbf{On the Robustness of SAM}}

Despite its generality to open-world imaging scenarios, SAM is still vulnerable to some image perturbations. \cite{zhang2023attacksam} has investigated adversarial attacks on SAM. It shows that the SAM model is not robust against the attack of adversarial examples~\cite{szegedy2013intriguing,goodfellow2014explaining,kurakin2016adversarial}. Specifically, the basic goal of attack-SAM~\cite{zhang2023attacksam} is set to remove the masks, for which the authors design a CLIP-MSE loss. The results show that the model is very vulnerable under a white-box PGD attack~\cite{madry2017towards}. In the black-box setting, the model maintains robustness to some extent. On top of removing the original masks, the authors also experiment with generating new (target) masks, showing intriguing results. 
Additionally, \cite{huang2023robustness,wang2023empirical} also examine SAM's robustness against fifteen image corruptions of varying severity. They found that all corruptions, except blur-related ones, only slightly decreased (<5\%) pixel accuracy and intersection over union metrics in various datasets. Nevertheless, \cite{wang2023empirical} discovered that the decline in the above metric values in the perturbed image is larger in the segmentation task involving challenging objects, such as medical X-rays. Another recent work~\cite{qiao2023robustness} performs comprehensive evaluations on the robustness of SAM on corruption and beyond. Specifically, it interprets various corruptions as new styles and tests SAM's robustness against style transfer and common corruptions. Moreover, it investigates SAM's robustness against local occlusion and adversarial perturbation. The results demonstrate that SAM has a moderate level of resilience against FGSM attacks, but not PGD attacks, even for perturbation with a very small magnitude~\cite{qiao2023robustness}.

\subsection{Future Directions}

In this subsection, we suggest potential strategies to boost the performance of both SAM and SAM 2. These strategies include fine-tuning, integration with other AI technologies, and architectural modifications, such as adapter fixing and memory optimization.

\subsubsection{\textbf{Improvement by Fine-Tuning}}

Fine-tuning is the quickest way to improve model's performance in the non-performing domain.
This strategy is mostly used in out-of-distribution segmentation, such as in medical imaging.
To this end, such fine-tuning approaches have been evaluated on polyp colonoscopy~\cite{li2023polypsam}, skin lesions~\cite{hu2023skinsam}, and other medical image datasets~\cite{ma2023segment}. On average, this strategy yields segmentation accuracy (Dice score) above 80\%. 
Ma \textit{et al.}~\cite{ma2023segment}, presented a fine-tuned SAM model called \textit{MedSAM}, trained on a large-scale medical image dataset with over 200,000 masks across 11 different modalities. Their approach outperformed Vanilla SAM with an average Dice similarity coefficient of 22.5\% on 3D images and 17.6\% on 2D images. 
Later, Li \textit{et al.}\cite{li2023polypsam} fine-tuned SAM through transfer learning on a colonoscopy image dataset. They found that either by training only the mask decoder or all the modules, their procedure yields segmentation results that outperform existing methods. Similarly, Hu \textit{et al.}\cite{hu2023skinsam} also conducted fine-tuning of SAM on a skin lesions dataset and found that domain-specific fine-tuning improves the segmentation performance of SAM.
Moreover, SAM 2 further benefits from fine-tuning by accommodating a broader range of medical image types and modalities. This makes it a more adaptable tool for complex medical segmentation tasks~\cite{zhu2024medical}.

\subsubsection{\textbf{Adapter Fixing}}

More advanced accuracy improvement approaches are proposed by slightly modifying SAM's framework. Most of these modifications are performed by attaching domain-specific adapters, whose role is to learn task-specific knowledge. Such an adapter can be fixed between transformer layers of the image encoder~\cite{qiu2023learnable, zhang2023customized, wu2023medical}, between attention layers of the mask decoder~\cite{wu2023medical}, or even completely replacing both the prompt encoder and mask decoder with a task-specific head~\cite{qiu2023learnable}. Another modification is proposed by decoupling the mask decoder into two modules which are responsible for handling IoU regression and mask learning respectively~\cite{gao2023desam}. Such a mechanism attempts to solve the coupling effect between image embedding and prompt token in the mask decoder that makes SAM segmentation output highly dependent on the prompt quality.

In the ophthalmology field, \cite{qiu2023learnable} inserts SAM's image encoder with task-specific prompt layers between transformer layers, which intend to learn task-specific knowledge. SAM's prompt generator and mask decoder are also replaced with a task-specific head. Although it still fails against minuscule objects, this approach outperforms other models in most datasets. \cite{zhang2023customized} also executed a similar task-specific insertion approach in SAM's image decoder layer. However, they kept both the prompt encoder and mask decoder while using default embedding at the same time, eliminating the need for prompting. Moreover, \cite{wu2023medical} inserts two learnable adapters at each encoder layer, specifically after the multi-head attention and at the residual connection. In the mask decoder module, three more adapters are also inserted after token-image attention, MLP, and image-token attention respectively. \cite{gao2023desam} argue that SAM's suboptimal performance in medical image segmentation is caused by the coupling effect between image embedding and prompt token in the mask decoder. Hence, they suggest decoupling this decoder into two which are responsible for IoU regression and mask learning respectively.

Furthermore, a recent study by Chen \textit{et al.}~\cite{chen2024sam2} demonstrates that using a specialized adapter strategy outperforms SAM 2. This success underscores the effectiveness of adapter fixing in improving segmentation performance for specific applications.




\subsubsection{\textbf{Integration with Other AI Technologies}}

The integration of SAM models with other AI technologies presents a fertile ground for enhancing segmentation capabilities. SAM 2, with its improvements in video and image segmentation, still relies heavily on prompts to achieve optimal performance. Integrating SAM with Large Language Models (LLMs) like GPT-4 or BERT~\cite{devlin2018bert} could enable more sophisticated interpretation of complex visual scenes by leveraging natural language understanding. This integration can facilitate more accurate prompt generation and refinement, allowing SAM to handle ambiguous or cluttered environments with greater precision.

Multimodal AI Systems offer another avenue for advancement. By combining SAM with audio and spatial data processing systems, applications can be developed in areas such as autonomous vehicles and interactive AI environments. For instance, in augmented reality (AR) and virtual reality (VR), SAM can be used to create real-time, adaptive environments that respond to both visual and auditory inputs, enhancing user experience through more immersive interactions.

In autonomous driving, the fusion of SAM with LIDAR and radar technologies could improve the model's ability to recognize and respond to dynamic and unpredictable road conditions. By integrating various sensor data, SAM can achieve a more comprehensive understanding of its surroundings, leading to safer and more reliable autonomous systems.

\subsubsection{\textbf{Extending Capabilities to New Domains and Datasets}}

While SAM has demonstrated strong performance in many segmentation tasks, there are significant opportunities to extend its capabilities to new and challenging domains. For example, SAM's performance in camouflaged and transparent object detection is an area that requires further exploration and enhancement~\cite{zhu2024medical}. The model's ability to accurately segment objects that blend into their backgrounds or are transparent is crucial for applications in surveillance, security, and wildlife monitoring.

In non-Euclidean data domains, such as graph-based structures, SAM could be adapted to segment data in fields like social network analysis and molecular biology. This would involve developing algorithms capable of handling irregular and heterogeneous graph structures, broadening SAM's applicability beyond traditional image data~\cite{yu2024sam}.

Furthermore, SAM models could be employed in satellite imagery analysis, where segmenting vast and diverse geographical data is essential for environmental monitoring, urban planning, and disaster management. In medical diagnostics, extending SAM's capabilities to handle a wider variety of medical imaging modalities and conditions could significantly impact early disease detection and personalized treatment planning~\cite{li2024adapting}.

\subsubsection{\textbf{Potential Improvements in Memory and Propagation Mechanisms}}

Enhancing the memory and propagation mechanisms within SAM models is crucial for improving their scalability and performance, especially in resource-constrained environments. SAM 2 introduces a memory bank that retains past predictions,  facilitating better video segmentation~\cite{rafaeli2024prompt, geetha2024sam}. However, the high computational costs, especially from the self-attention for frame embedding and cross-attention to past memory features, call for more sophisticated memory architectures and processing techniques. These optimizations could significantly reduce the overhead associated with large-scale segmentation tasks, and can further enhance the model's efficiency and accuracy. 

\begin{itemize}

\item \textbf{Memory Management.} Researchers could explore advanced memory architectures that allow for more efficient storage and retrieval of information, reducing the computational overhead associated with large-scale segmentation tasks. Techniques like memory-efficient attention mechanisms could be employed to streamline the processing of visual information without sacrificing accuracy.

\item \textbf{Propagation Techniques.} Improving how information is propagated across frames or slices is particularly relevant for 3D medical imaging and video segmentation~\cite{yu2024novel}. Developing adaptive propagation methods that consider spatial and temporal variations can lead to more accurate segmentation across complex datasets~\cite{zhu2024medical}.

\item \textbf{Unsupervised and Reinforcement Learning.} Incorporating unsupervised learning techniques could reduce SAM's dependency on extensive labeled datasets, allowing it to learn more generalizable features across diverse tasks and datasets. Reinforcement learning can be employed to refine segmentation strategies dynamically, adapting to new environments and challenges without requiring manual intervention.

\end{itemize}

\noindent The future of the SAM family lies in its ability to integrate with complementary AI technologies, adapt to new domains, and refine its internal mechanisms to handle complex segmentation tasks more effectively. By addressing the current limitations and exploring these promising directions, researchers can unlock the full potential of SAM models, driving innovation across various fields and applications. Continued research and development in these areas will ensure that SAM remains at the forefront of segmentation technology, offering solutions that are both versatile and robust.

\section{Conclusion}

This survey presents the first comprehensive review of the Segment Anything Model (SAM) family, encompassing both SAM and SAM 2, and their transformative impact on the field of computer vision. By exploring the development, innovations, and applications of SAM, we highlight its groundbreaking approach to segmentation across diverse tasks and data types.
Our work thoroughly examines SAM's capabilities and limitations, revealing its potential to revolutionize a wide array of applications, from medical diagnostics to industrial automation. Despite its impressive versatility, SAM still faces challenges in scenarios requiring high granularity and contextual understanding, particularly where explicit prompts are unavailable.
This survey provides valuable insights into the strengths and weaknesses of SAM, offering guidance for future research aimed at enhancing its robustness and generalization capabilities. By identifying key areas for improvement and potential applications, we believe our work serves as a foundational resource for advancing SAM technologies and exploring new frontiers in computer vision. As SAM continues to evolve, ongoing research and collaboration will be essential in unlocking its full potential and addressing the complex challenges of segmentation in ever-changing environments.


\bibliographystyle{IEEEtran}

\bibliography{bib_sam}

\end{document}